# Flying Insect Classification with Inexpensive Sensors

Yanping Chen

*Department of Computer Science & Engineering,*
*University of California, Riverside*

ychen053@ucr.edu

Adena Why

*Department of Entomology,*
*University of California, Riverside*

awhy001@ucr.edu

Gustavo Batista

*University of São Paulo - USP*

gbatista@icmc.usp.br

Agenor Mafra-Neto

*ISCA Technologies*

president@iscatech.com

Eamonn Keogh

*Department of Computer Science & Engineering,*
*University of California, Riverside*

eamonn@cs.ucr.edu

## Abstract

The ability to use inexpensive, noninvasive sensors to accurately classify flying insects would have significant implications for entomological research, and allow for the development of many useful applications in vector control for both medical and agricultural entomology. Given this, the last sixty years have seen many research efforts on this task. To date, however, none of this research has had a lasting impact. In this work, we explain this lack of progress. We attribute the stagnation on this problem to several factors, including the use of *acoustic* sensing devices, the overreliance on the single feature of wingbeat frequency, and the attempts to learn complex models with relatively little data. In contrast, we show that pseudo-acoustic *optical* sensors can produce vastly superior data, that we can exploit additional features, both intrinsic and extrinsic to the insect's flight behavior, and that a Bayesian classification approach allows us to efficiently learn classification models that are very robust to overfitting. We demonstrate our findings with large scale experiments that dwarf all previous works combined, as measured by the number of insects and the number of species considered.

# Introduction

The idea of automatically classifying insects using the incidental sound of their flight (as opposed to *deliberate* insect sounds produced by stridulation (Hao et al. 2012)) dates back to the very dawn of computers and commercially available audio recording equipment. In 1945[1], three researchers at the Cornell University Medical College, Kahn, Celestin and Offenhauser, used equipment donated by Oliver E. Buckley (then President of the Bell Telephone Laboratories) to record and analyze mosquito sounds (Kahn et al. 1945).

The authors later wrote, "*It is the authors' considered opinion that the intensive application of such apparatus will make possible the precise, rapid, and simple observation of natural phenomena related to the sounds of disease-carrying mosquitoes and should lead to the more effective control of such mosquitoes and of the diseases that they transmit*." (Kahn and Offenhauser 1949). In retrospect, given the importance of insects in human affairs, it seems astonishing that more progress on this problem has not been made in the intervening decades.

[1] An even earlier paper (Reed et al. 1941) makes a similar suggestion. However, these authors determined the wingbeat frequencies *manually*, aided by a stroboscope.

There have been sporadic efforts at flying insect classification from audio features (Sawedal and Hall 1979; Schaefer and Bent 1984; Unwin and Ellington 1979; Moore et al. 1986**)**, especially in the last decade (Moore and Miller 2002; Repasky et al. 2006); however, little real progress seems to have been made. By "lack of progress" we do not mean to suggest that these pioneering research efforts have not been fruitful. However, we would like to have *automatic* classification to become as simple, inexpensive, and ubiquitous as current *mechanical* traps such as sticky traps or interception traps (Capinera 2008), but with all the advantages offered by a digital device: higher accuracy, very low cost, real-time monitoring ability, and the ability to collect additional information (time of capture[2], etc.).

We feel that the lack of progress in this pursuit can be attributed to three related factors:

1. Most efforts to collect data have used *acoustic* microphones (Reed et al. 1942; Belton et al. 1979; Mankin et al. 2006; Raman et al. 2007). Sound attenuates according to an inverse squared law. For example, if an insect flies just three times further away from the microphone, the *sound intensity* (informally, the *loudness*) drops to one ninth. Any attempt to mitigate this by using a more sensitive microphone invariably results in extreme sensitivity to wind noise and to ambient noise in the environment. Moreover, the difficulty of collecting data with such devices seems to have led some researchers to obtain data in unnatural conditions. For example, nocturnal insects have been forced to fly by tapping and prodding them under bright halogen lights; insects have been recorded in confined spaces or under extreme temperatures (Belton et al. 1979; Moore and Miller 2002). In some cases, insects were tethered with string to confine them within the range of the microphone (Reed et al. 1942). It is hard to imagine that such insect handling could result in data which would generalize to insects in natural conditions.
2. Unsurprisingly, the difficultly of obtaining data noted above has meant that many researchers have attempted to build classification models with very limited data, as few as 300 instances (Moore 1991) or less. However, it is known that for building classification models, more data is better (Halevy et al. 2009; Banko and Brill 2001; Shotton et al. 2013).

[2] A commercially available rotator bottle trap made by BioQuip® (2850) does allow researchers to measure the time of arrival at a granularity of *hours*. However, as we shall show in Section *Additional Feature: Circadian Rhythm of Flight Activity*, we can measure the time of arrival at a *sub-second* granularity and exploit this to improve classification accuracy.

3. Compounding the poor quality data issue and the sparse data issue above is the fact that many researchers have attempted to learn very complicated classification models[3], especially neural networks (Moore et al. 1986; Moore and Miller 2002; Li et al. 2009). However, neural networks have many parameters/settings, including the interconnection pattern between different layers of neurons, the learning process for updating the weights of the interconnections, the activation function that converts a neuron's weighted input to its output activation, etc. Learning these on say a spam/email classification problem with millions of training data is not very difficult (Zhan et al. 2005), but attempting to learn them on an insect classification problem with a mere twenty examples is a recipe for overfitting (cf. Figure 3). It is difficult to overstate how optimistic the results of neural network experiments can be unless rigorous protocols are followed (Prechelt 1995).

In this work, we will demonstrate that we have largely solved all these problems. We show that we can use optical sensors to record the "sound" of insect flight from meters away, with complete invariance to wind noise and ambient sounds. We demonstrate that these sensors have allowed us to record on the order of *millions* of labeled training instances, far more data than all previous efforts combined, and thus allow us to avoid the overfitting that has plagued previous research efforts. We introduce a principled method to incorporate additional information into the classification model. This additional information can be as quotidian and as easy-to-obtain as the time-of-day, yet still produce significant gains in accuracy. Finally, we demonstrate that the enormous amounts of data we collected allow us to take advantage of "*The unreasonable effectiveness of data*" (Halevy et al. 2009) to produce simple, accurate and robust classifiers.

In summary, we believe that flying insect classification has moved beyond the dubious claims created in the research lab and is now ready for real-world deployment. The sensors and software we present in this work will provide researchers worldwide robust tools to accelerate their research.

[3] While there is a formal framework to define the complexity of a classification model (i.e. the VC dimension (Vapnik and Chervonenkis 1971)), informally we can think of a *complicated* or *complex* model as one that requires many parameters to be set or learned.

## Background and Related Work

The vast majority of attempts to classify insects by their flight sounds have explicitly or implicitly used *just* the wingbeat frequency (Reed et al. 1942; Sotavalta 1947; Sawedal and Hall 1979; Schaefer and Bent 1984; Unwin and Ellington 1979; Moore et al. 1986; Moore 1991). However, such an approach is limited to applications in which the insects to be discriminated have very different frequencies. Consider Figure 1.I which shows a histogram created from measuring the wingbeat frequencies of three (sexed) species of insects, *Culex stigmatosoma* (female), *Aedes aegypti* (female), and *Culex tarsalis* (male) (We defer details of how the data was collected until later in the paper).

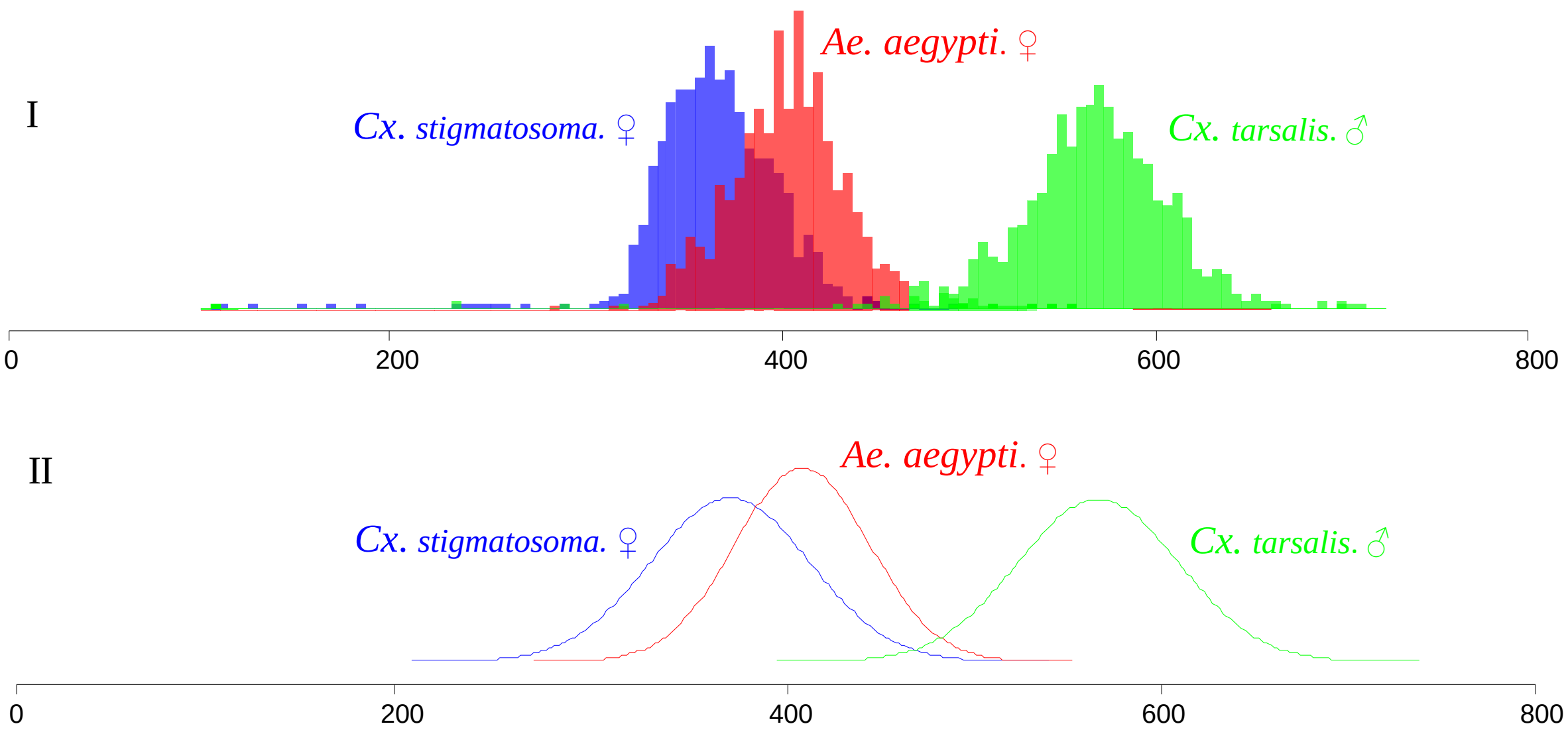

**Figure 1: I) Histograms of wingbeat frequencies of three species of insects, *Cx. stigmatosoma* ♀, *Ae. aegypti.* ♀, and *Cx. tarsalis.* ♂. Each histogram is derived based on 1,000 wingbeat sound snippets. II) Gaussian curves that fit the wingbeat frequency histograms**

It is visually obvious that if asked to separate *Cx. stigmatosoma* ♀ from *Cx. tarsalis* ♂, the wingbeat frequency could produce an accurate classification, as the two species have very different frequencies with minimal overlap. To see this, we can compute the optimal Bayes error rate (Fukunaga 1990), which is a strict lower bound to the actual error rate obtained by *any* classifier that considers *only* this feature. Here, the Bayes error rate is half the *overlapping* area under both curves divided by the *total* area under the two curves.

Because there is only a tiny overlap between the wingbeat frequency distributions of the two species, the Bayes error rate is correspondingly small, 0.57% if we use the raw histograms and 1.08% if we use the derived Gaussians.

However, if the task is to separate *Cx. stigmatosoma.* ♀ from *Ae. aegypti.* ♀, the wingbeat frequency will not do as well, as the frequencies of these two species overlap greatly. In this case, the Bayes error rate is *much* larger, 24.90% if we use the raw histograms and 30.95% if we use the derived Gaussians.

This problem can only get worse if we consider more species, as there will be increasing overlap among the wingbeat frequencies. This phenomenon can be understood as a real-value version of the Pigeonhole principle (Grimaldi 1989). As a concrete example, assume we limit our attention to just mosquitoes. There are more than 3,500 species of mosquitoes described worldwide. Assume the range of the mosquito wingbeat frequency is between 100 Hz and 1,000 Hz. If each species takes up an integer wingbeat frequency, then at least 2,600 species must share a same wingbeat frequency with another species, as there are only 900 possible frequency values. The actual overlap rate will be even higher because the typical wingbeat frequency for a species is a distribution peaking at some value rather than just a single integer value, as shown in Figure 1.I.

Given this, it is unsurprising that some doubt the utility of wingbeat sounds to classify the insects. However, we will show that the analysis above is pessimistic. Insect flight sounds can allow much higher classification rates than the above suggests because:

- There is more information in the flight sound signal than just the wingbeat frequency. By analogy, humans have no problem distinguishing between Middle C on a piano and Middle C on a saxophone, even though both are the same 261.62 Hz fundamental frequency. The Bayes error rate to classify the three species in Figure 1.I using *just* the wingbeat frequency is 19.13%; however, as we shall see below in the section titled *Flying Insect Classification*, that by using the additional features from the signal, we can obtain an error rate of 12.43%. We believe that our experiments are the first explicit demonstration that there is actionable information in the signal beyond the wingbeat frequency.

- We can augment the wingbeat sounds with additional cheap-to-obtain features that can help to improve the classification performance. For example, many species may have different flight activity circadian rhythms. As we shall see below in the section titled *Additional Feature: Circadian Rhythm of Flight Activity*, simply incorporating the *time-of-intercept* information can significantly improve the performance of the classification.

The ability to allow the incorporation of auxiliary features is one of the reasons we argue that the Bayesian classifier is ideal for this task (cf. Section *Flying Insect Classification*), as it can gracefully incorporate evidence from multiple sources and in multiple formats. One of the meta-features that the Bayesian classifier can incorporate is the prior probability of seeing a particular insect. In some cases, we may be able to further improve the accuracy of classification by adjusting these prior probabilities with some on-site intervention. For example, we may use attractants or repellants, or we may construct the sensors with mechanical barriers that limit entry for large insects, or fans that discourage weak flyers, etc. We leave such considerations to future work.

# Materials and Methods

## Insect Colony and Rearing

Six species of insects were studied in this work: *Cx. tarsalis, Cx. stigmatosoma*, *Ae. aegypti*, *Culex quinquefasciatus*, *Musca domestica* and *Drosophila simulans*.

All adult insects were reared from laboratory colonies derived from wild individuals collected at various locations. *Cx. tarsalis* colony was derived from wild individuals collected at the Eastern Municipal Water District's demonstration constructed treatment wetland (San Jacinto, CA) in 2001. *Cx. quinquefasciatus* colony was derived from wild individuals collected in southern California in 1990 (Georghiou and Wirth 1997). *Cx. stigmatosoma* colony was derived from wild individuals collected at the University of California, Riverside, Aquatic Research Facility in Riverside, CA in 2012. *Ae. aegypti* colony was started in 2000 with eggs from Thailand (Van Dam and Walton 2008). *Musca domestica* colony was derived from wild individuals collected in San Jacinto, CA in 2009, and *Drosophila simulans* colony were derived from wild individuals caught in Riverside, CA in 2011.

The larvae of *Cx. tarsalis, Cx. quinquefasciatus, Cx. stigmatosoma* and *Ae. aegypti* were reared in enamel pans under standard laboratory conditions (27 °C, 16:8 h light:dark [LD] cycle with 1 hour dusk/dawn periods) and fed ad libitum on a mixture of ground rodent chow and Brewer's yeast (3:1, v:v). *Musca domestica* larvae were kept under standard laboratory conditions (12:12 h light:dark [LD] cycle, 26 °C, 40% RH) and reared in a mixture of water, bran meal, alfalfa, yeast, and powdered milk. *Drosophila simulans* larvae were fed ad libitum on a mixture of rotting fruit.

Mosquito pupae were collected into 300-mL cups (Solo Cup Co., Chicago IL) and placed into experimental chambers. Alternatively, adults were aspirated into experimental chambers within 1 week of emergence. The adult mosquitoes were allowed to feed ad libitum on a 10% sucrose and water mixture; food was replaced weekly. Cotton towels were moistened, twice a week, and placed on top of the experimental chambers and a 300-ml cup of tap water (Solo Cup Co., Chicago IL) was kept in the chamber at all times to maintain a higher level of humidity within the cage. *Musca domestica* adults were fed ad libitum on a mixture of sugar and low-fat dried milk, with free access to water. *Drosophila simulans* adults were fed ad libitum on a mixture of rotting fruit.

Experimental chambers consisted of Kritter Keepers (Lee's Aquarium and Pet Products, San Marcos, CA) that were modified to include the sensor apparatus as well as a sleeve (Bug Dorm sleeve, Bioquip, Rancho Dominguez, CA) attached to a piece of PVC piping to allow access to the insects. Two different sizes of experimental chambers were used, the larger 67 cm L x 22 cm W x 24.75 cm H, and the smaller 30 cm L x 20 cm W x 20 cm H. The lids of the experimental chambers were modified with a piece of mesh cloth affixed to the inside in order to prevent escape of the insects, as shown in Figure 2.I. Experimental chambers were maintained on a 16:8 h light:dark [LD] cycle, 20.5-22 °C and 30-50% RH for the duration of the experiment. Each experimental chamber contained 20 to 40 individuals of a same species, in order to capture as many flying sounds as possible while limiting the possibility of capturing more than one insect-generated sound at a same time.

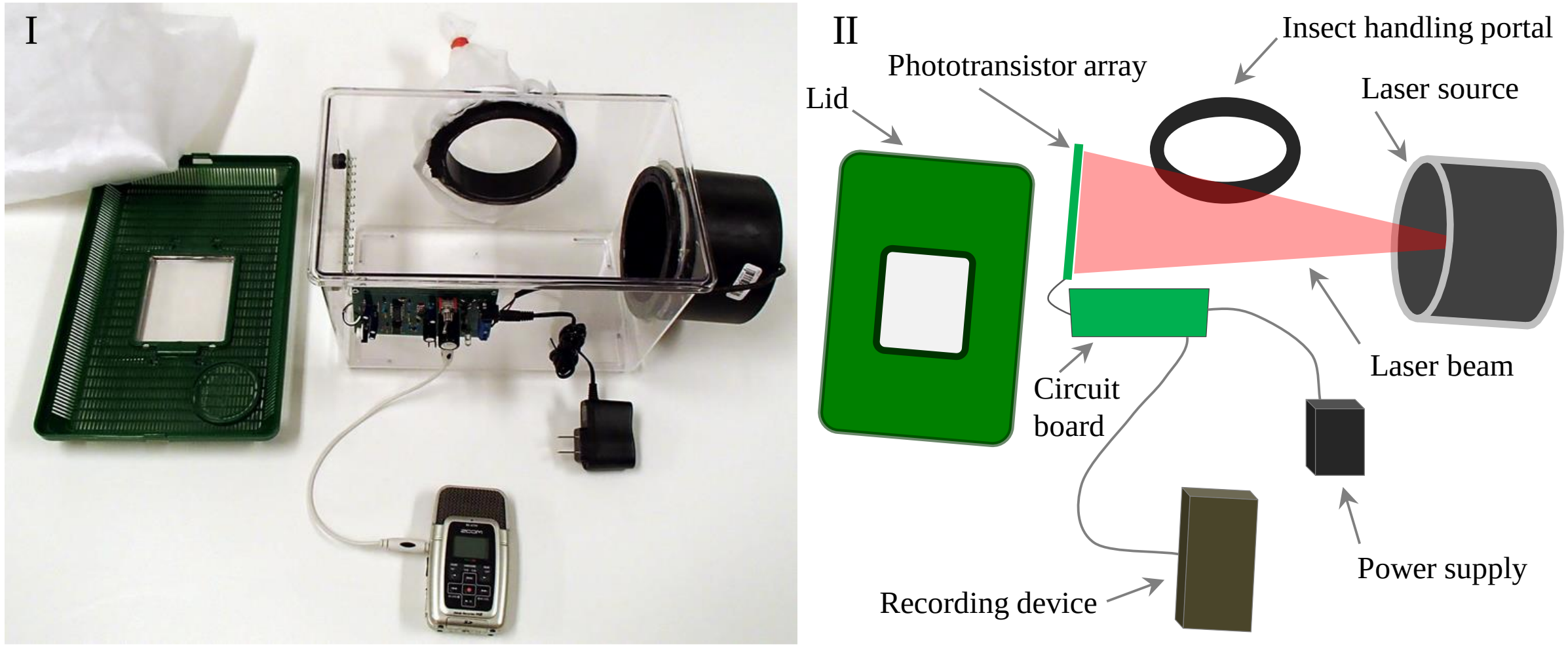

**Figure 2: I) One of the cages used to gather data for this project. II) A logical version of the setup with the components annotated**

### Instruments to Record Flying Sounds

We used the sensor described in (Batista 2011) to capture the insect flying sounds. The logic design of the sensor consists of a phototransistor array which is connected to an electronic board, and a laser line pointing at the phototransistor array. When an insect flies across the laser beam, its wings partially occlude the light, causing small light fluctuations. The light fluctuations are captured by the phototransistor array as changes in current, and the signal is filtered and amplified by the custom designed electronic board. The physical version of the sensor is shown in Figure 2.I.

The output of the electronic board feeds into a digital sound recorder (Zoom H2 Handy Recorder) and is recorded as audio data in the MP3 format. Each MP3 file is 6 hours long, and a new file starts recording immediately after a file has recorded for 6 hours, so the data is continuous. The length of the MP3 file is limited by the device firmware rather than the disk space. The MP3 standard is a lossy format and optimized for human perception of speech and music. However, most flying insects produce sounds that are well within the range of human hearing and careful comparisons to lossless recordings suggest that we lose no exploitable (or indeed, *detectable*) information.

## Sensor Data Processing

We downloaded the MP3 sound files to a PC twice a week and used a detection algorithm to automatically extract the brief insect flight sounds from the raw recording data. The detection algorithm used a sliding window to "slide" through the raw data. At each data point, a classifier/detector is used to decide whether the audio segment contains an insect flying sound. It is important to note that the classifier used at this stage is solving the relatively simple two-class task, differentiating between *insect|non-insect*. We will discuss the more sophisticated classifier, which attempts to differentiate species and sex, in the next section.

The classifier/detector used for the *insect|non-insect* problem is a nearest neighbor classifier based on the frequency spectrum. For ground truth data, we used ten flying sounds extracted from early experiments as the training data for the *insect* sounds, and ten segments of raw recording background noise as the training data for the *non-insect* sounds. The number of training data was limited to ten, because more training data would slow down the algorithm while fewer data would not represent variability observed. Note that the training data for background sounds can be different from minute to minute. This is because while the frequency spectrum of the background sound has little variance within a short time interval, it can change greatly and unpredictably in the long run. This variability (called *concept drift* in the machine learning community (Tsymbal 2004; Widmer and Kubat 1996)) may be due to the effects of temperature change on the electronics and the slow decline of battery output power etc. Fortunately, given the high signal-to-noise ratio in the audio, the high variation of the *non-insect* sounds does not cause a significant problem. Figure 5.I shows an example of a one-second audio clip containing a flying insect generated by our sensor. As we can see, the signal of insects flying across the laser is well distinguished from the background signal, as the amplitude is much higher and the range of frequency is quite different from that of background sound.

The length of the sliding window in the detection algorithm was set to be 100 ms, which is about the average length of a flying sound. Each detected insect sound is saved into a one-second long WAV format audio file by centering the insect flying signal and padding with zeros elsewhere. This makes all flying sounds the same length and simplifies the future archiving and processing of the data. Note that we converted the audio format from MP3 to WAV at this stage. This is simply because we publicly release all our data so that the community can confirm and extend our results. Because the vast majority of the signal processing community uses Matlab, and Matlab provides native functions for working with WAV files, this is the obvious choice for an archiving format. Figure 5.II shows the saved audio of the insect sound shown in Figure 5.I.

Flying sounds detected in the raw recordings may be contaminated by the background noise, such as the 60 Hz noise from the American domestic electricity, which “bleeds” into the recording due to the inadequate filtering in power transformers. To obtain a cleaner signal, we applied the spectral subtraction technique (Boll 1979; Ephraim and Malah 1984) to each detected flying sound to reduce noise.

## Flying Insect Classification

In the section above, we showed how a simple nearest neighbor classifier can *detect* the sound of insects, and pass the sound snippet on for further inspection. Here, we discuss algorithms to actually classify the snippets down to species (and in some cases, sex) level.

While there are a host of classification algorithms in the literature (decision trees, neural networks, nearest neighbor, etc.), the Bayes classifier is optimal in minimizing the probability of misclassification (Devroye 1996), under the assumption of independence of features. The Bayes classifier is a simple probabilistic classifier that predicts class membership probabilities based on Bayes’ theorem. In addition to its excellent classification performance, the Bayesian classifier has several properties that make it extremely useful in practice and particularly suitable to the task at hand.

1. The Bayes classifier is undemanding in both CPU and memory requirements. Any devices to be deployed in the field in large quantities will typically be small devices with limited resources, such as limited memory, CPU power and battery life. The Bayesian classifier (once constructed offline in the lab) requires time and space resources that are just linear in the number of features.
2. The Bayes classifier is very easy to implement. Unlike neural networks (Moore and Miller 2002; Li et al. 2009), the Bayes classifier does not have many parameters that must be carefully tuned. In addition, the model is fast to build, and it requires only a small amount of training data to estimate the distribution parameters necessary for accurate classification, such as the means and variances of Gaussian distributions.

3. Unlike other classification methods that are essentially "black box", the Bayesian classifier allows for the graceful introduction of user knowledge. For example, if we have external (to the training data set) knowledge that given the particular location of a deployed insect sensor we should expect to be twice as likely to encounter a *Cx. tarsalis* as an *Ae. aegypti*, we can "tell" the algorithm this, and the algorithm can use this information to improve its accuracy. This means that in some cases, we can augment our classifier with information gleaned from the text of journal papers or simply the experiences of field technicians. In Section *A Tentative Additional Feature: Geographic Distribution*, we give a concrete example of this.

4. The Bayesian classifier simplifies the task flagging anomalies. Most classifiers must make a classification decision, even if the object being classified is vastly different to anything observed in the training phase. In contrast, we can slightly modify the Bayesian classifier to produce an "*Unknown*" classification. One or two such classifications per day could be ignored, but a spate of them could be investigated in case it is indicative of an infestation of a completely unexpected invasive species.

5. When there are multiple features used for classification, we need to consider the possibility of missing values, which happens when some features are not observed. For example, as we discuss below, we use *time-of-intercept* as a feature. However, a dead clock battery could deny us this feature even when the rest of the system is working perfectly. Missing values are a problem for any learner and may cause serious difficulties. However, the Bayesian classifier can trivially handle this problem, simply by dynamically ignoring the feature in question at classification time.

Because of the considerations listed above, we argue that the Bayesian classifier is the best for our problem at hand. Note that our decision to use Bayesian classifier, while informed by the above advantages, was also informed by an extensive empirical comparison of the accuracy achievable by other methods, given that in some situations *accuracy* trumps all other considerations. While we omit exhaustive results for brevity, in Figure 3 we show a comparison with the neural network classifier, as it is the most frequently used technique in the literature (Moore and Miller 2002). We considered only the frequency spectrum of wingbeat snippets for the three species discussed in Figure 1. The training data was randomly sampled from a pool of 1,500 objects, and the test data was a completely disjoint set of 1,500 objects, and we tested over 1,000 random resamplings. For the neural network, we used a single hidden layer of size ten, which seemed to be approximately the default parameters in the literature.

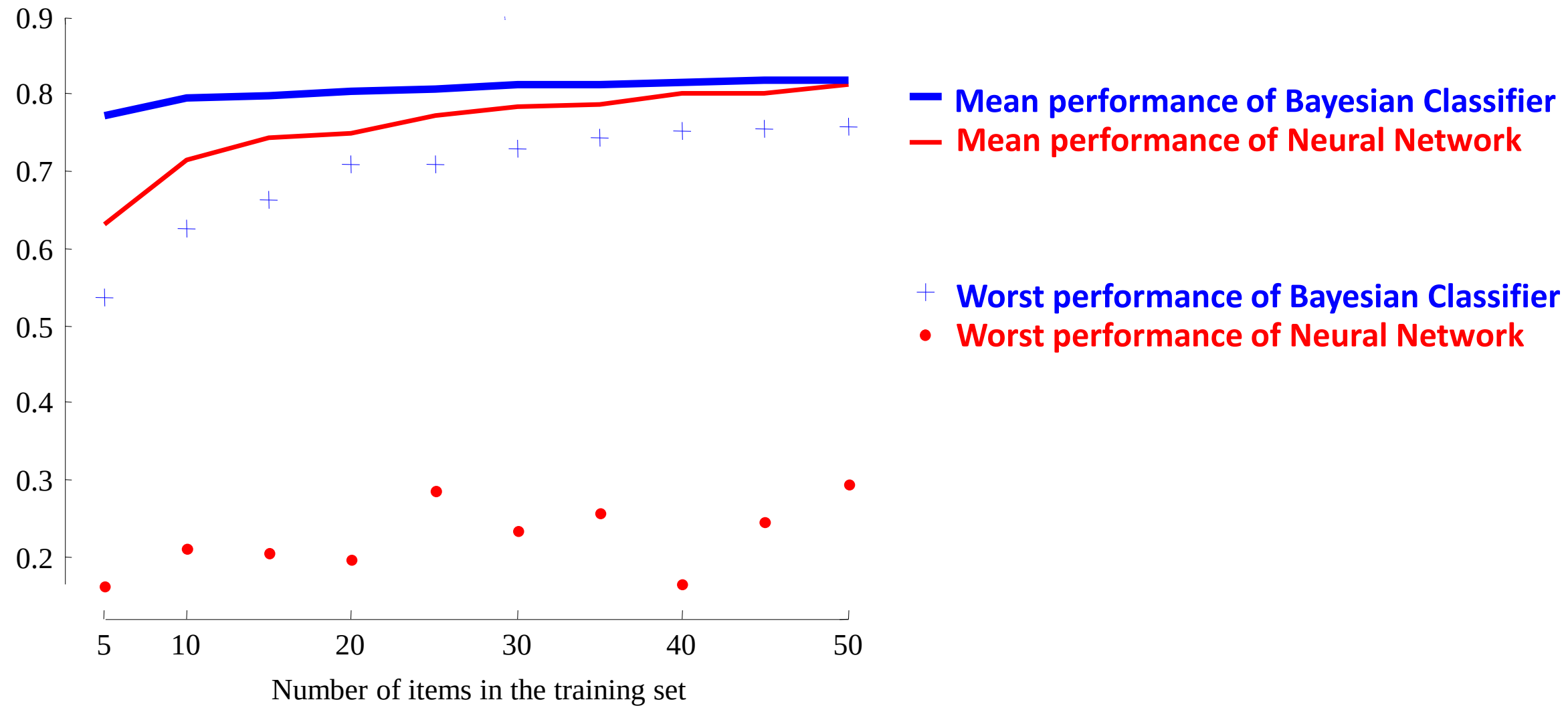

**Figure 3: A comparison of the mean and worst performance of the Bayesian versus Neural Networks Classifiers for datasets ranging in size from five to fifty.**

The results show that while the neural network classifier eventually converges on the performance of the Bayesian classifier, it is significantly worse for smaller datasets. Moreover, for any dataset size in the range examined, it can occasionally produce pathologically poor results, doing worse than the default rate of 33.3%.

Note that our concern about performance on small datasets is only apparently in conflict with our claim that our sensors can produce massive datasets. In some cases, when dealing with new insect species, it may be necessary to bootstrap the modeling of the species by using just a handful of annotated examples to find more (unannotated) examples in the archives, a process known as semi-supervised learning (Chen et al. 2013).

The intuition behind Bayesian classification is to find the mostly likely class given the data observed. The probability that an observed data $X$ belongs to a class $C_i$, $P(C_i|X)$, is computed using the Bayes rule as:

$$P(C_{\mathrm{i}}|X) = \frac{P(C_i)P(X|C_i)}{P(X)}$$

where $P(C_{\mathrm{i}})$ is a *prior* probability of class $C_{\mathrm{i}}$, that can be estimated from frequencies in the database; $P(X|C_i)$ is the probability of observing the data $X$ in class $C_{\mathrm{i}}$; and $P(X)$ is the probability of occurrence of the observed data $X$. The probability $P(X)$ is usually unknown, but since it does not depend on the class, it is usually understood as a normalization factor, and thus, only the numerator is considered for classification. The probability $P(C_{\mathrm{i}}|X)$ is proportional to the numerator:

$$P(C_{\mathrm{i}}|X) \propto P(C_i)P(X|C_i) \tag{1}$$

The $P(C_{\mathrm{i}}|X)$ is called the *posterior* probability. The Bayesian classifier assigns the data to the class $\hat{C}$ which has the highest *posterior* probability, that is,

$$\hat{C} = \underset{C_i \in C}{\operatorname{argmax}} P(C_i|X) \tag{2}$$

Where $C$ is the set of classes, i.e. {$C_1$ = *Cx. stigmatosoma*, $C_2$ = *Ae. aegypti*, ... ,$C_n$ = *An. gambiae*}

A Bayesian classifier can be represented using a graph called Bayesian network. The Bayesian network that uses a single feature for classification is shown in Figure 4. The direction of the arrow in the graph encodes the fact that the probability of an insect to be a member of class *C* depends on the value of the feature $F_1$ observed (i.e., *wingbeat-frequency*, *time-of-intercept, etc*).

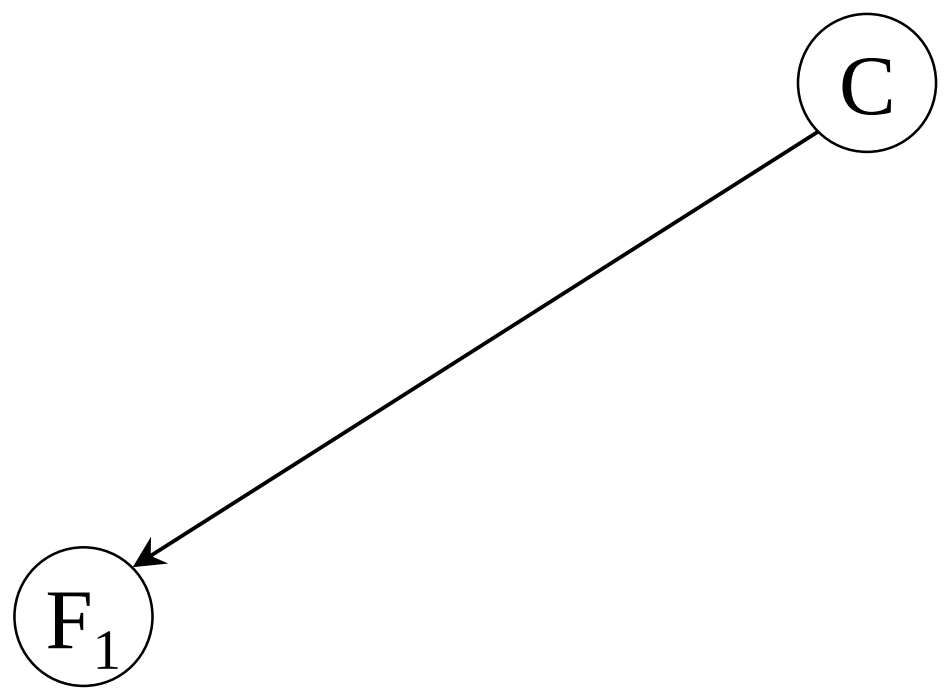

**Figure 4: A Bayesian network that uses a single feature for classification**

When the classifier is based on a single feature $F_1$, the *posterior* probability that an observed data $f_1$ belongs to a class $C_i$ is calculated as:

$$P(C_i|\mathrm{F}_1 = f_1) \propto P(C_i)P(\mathrm{F}_1 = f_1|C_i) \quad (3)$$

Where $P(\mathrm{F}_1 = f_1|C_i)$ is the class-conditioned probability of observing feature $f_1$ in class $C_i$.

For insect classification, the primary data we observed are the flight sounds, as illustrated in Figure 5.I. The flying sound signal is the non-zero amplitude section (red/bold) in the center of the audio, and can be represented by a sequence S = $<s_1, s_2, \ldots s_N>$, where $s_i$ is the signal sampled in the instance $i$ and $N$ is the total number of samples of the signal. This sequence contains a lot of acoustic information, and features can be extracted from it.

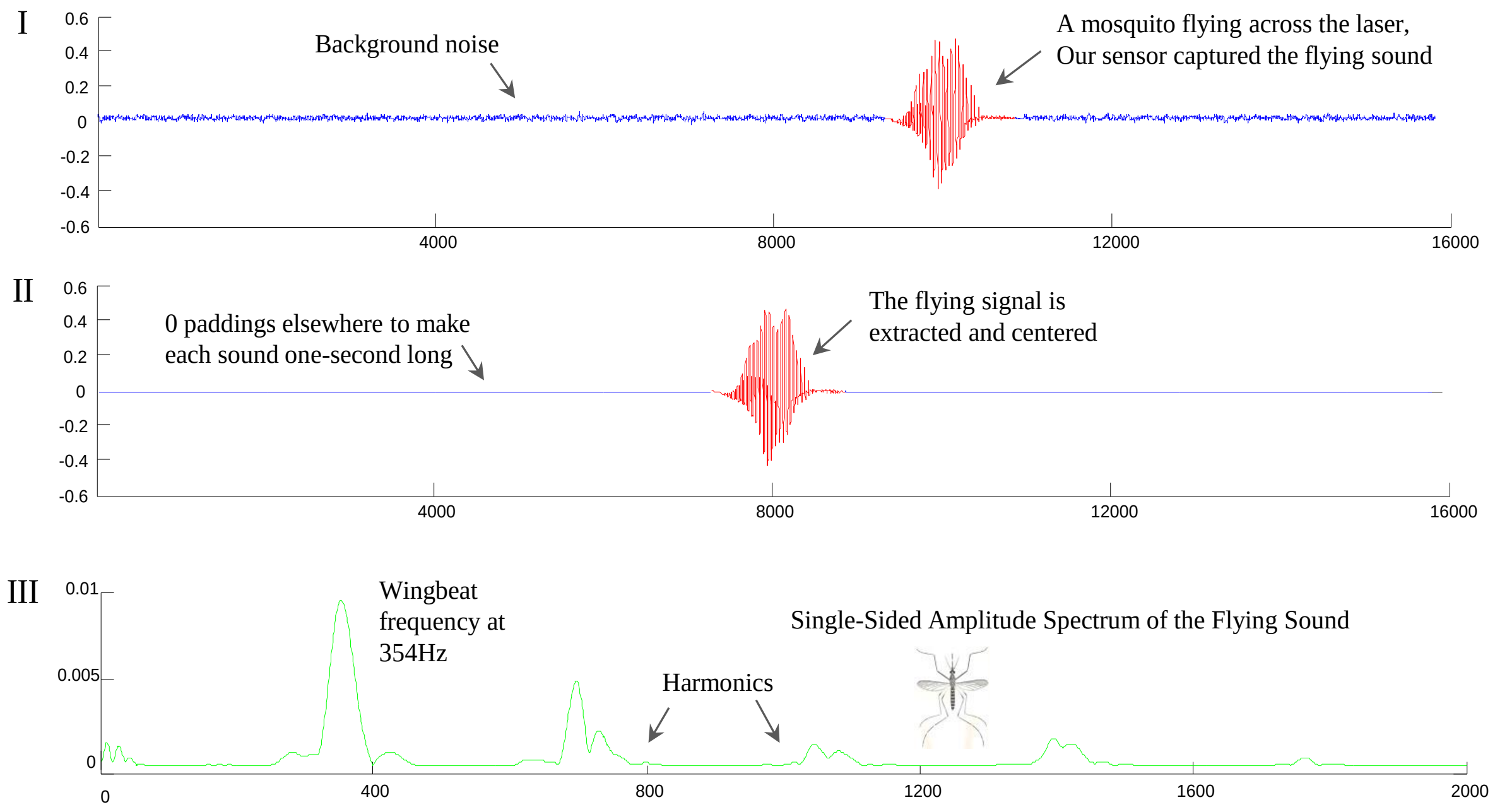

**Figure 5: I) An example of a one-second audio clip containing a flying sound generated by the sensor. The sound was produced by a female *Cx. stigmatosoma*. The insect sound is highlighted in red/bold. II) The insect sound that is cleaned and saved into a one-second long audio clip by centering the insect signal and padding with 0s elsewhere. III) The frequency spectrum of the insect sound obtained using DFT**

The most obvious feature to extract from the sound snippet is the wingbeat frequency. To compute the wingbeat frequency, we first transform the audio signal into a frequency spectrum using the Discrete Fourier Transform (DFT) (Bracewell and Bracewell 1986). As shown in Figure 5.III, the frequency spectrum of the sound in Figure 5.II has a peak in the fundamental frequency at 354 Hz, and some harmonics in integer multiples of the fundamental frequency. The highest peak represents the frequency of interest, i.e., the insect wingbeat frequency.

The class-conditioned wingbeat frequency distribution is a univariate density function that can be easily estimated using a histogram. Figure 1.I shows a wingbeat frequency histogram plot for three species of insects (each for a single sex only). We can observe that the histogram for each species is well modeled by Gaussian distribution. Hence, we fit a Gaussian for each distribution and estimated the means and variances using the frequency data. The fitted Gaussian distributions are shown in Figure 1.II. Note that as hinted at in the introduction, the Bayesian classifier does not have to use the idealized Gaussian distribution; it could use the raw histograms to estimate the probabilities instead. However, using the Gaussian distributions is computationally cheaper at classification time and helps guard against overfitting.

With these class-conditioned distributions, we can calculate the class-conditioned probability of observing the flying sound from a class, given the wingbeat frequency. For example, suppose the class of the insect shown in Figure 5.I is unknown, but we have measured its wingbeat frequency to be 354 Hz. Further suppose that we have previously measured the mean and standard deviation of female *Cx. stigmatosoma* wingbeat frequency to be 365 and 41, respectively. We can then calculate the probability of observing this wingbeat frequency from a female *Cx. stigmatosoma* insect using the Gaussian distribution function:

$$P(wingbeat = 354 \text{ Hz} | \text{female } Cx. stigmatosoma) = \frac{1}{41\sqrt{2\pi}} e^{-\frac{(354-365)^2}{2\times 41^2}}$$

We can calculate the probabilities for the other classes in a similar way, and predict the unknown insect as the most likely class using equation (2). In this example, the *prior* probability is equal for each class, and the unknown insect is about 2.1 times more likely to be a female *Cx. stigmatosoma* than be a female *Ae. aegypti* (the second most likely class), and thus is (in this case, *correctly*) classified as a female *Cx. stigmatosoma*.

Note that the wingbeat frequency is a scalar, and learning the class-conditioned density functions for a low-dimensional feature (typically no more than 3 dimensions) is easy, because they can either be fitted using some distribution models, such as the Gaussian distribution, or be approximated using a histogram which can be constructed with a small amount of training data. However, if a feature is high-dimensional, we need other multivariate density function estimation methods, because we usually do not have any idea of what distribution model should fit the distributions of the high-dimensional features, and building a histogram of a high-dimensional feature requires a prohibitively large size of training dataset (the size of training dataset grows exponentially with the increase of dimensionality).

The literature has offered some multivariate density function estimation methods for high-dimensional variables, such as the Parzen–Rosenblatt window method (Rosenblatt 1956; Parzen 1962) and the k-Nearest-Neighbors (kNN) approach (Mack and Rosenblatt 1979). The kNN approach is very simple; it leads to an approximation of the optimal Bayes classifier, and hence, we use it in this work to estimate the class-conditioned density functions for high-dimensional features.

The kNN approach does not require a training phase to learn the class-conditioned density function. It directly uses the training data to estimate the probability of observing an unknown data in a class. Specifically, given an observed data $X$, the kNN approach first searches the training data to find the top $k$ nearest neighbors of $X$. It then computes the probability of observing $X$ in class $C_{\mathrm{i}}$ as the fraction of the top $k$ nearest neighbors which are labeled as class $C_{\mathrm{i}}$, that is,

$$P(X|C_{\mathrm{i}}) = \frac{k_{C_i}}{k} \tag{4}$$

Where $k$ is a user-provided parameter specifying the number of nearest neighbors, and $k_{C_i}$ is the number of neighbors that are labeled as class $C_{\mathrm{i}}$ among the top $k$ nearest neighbors.

With equation (4), we can calculate the class-conditioned probability, and “plug” this into the Bayesian classifier. As an example, imagine that we use the entire spectrum as a feature for the insect sound (cf. Figure 5.II). Given an unknown insect, we first transform its flight sound into the spectrum representation, and then search the entire training data to find the top $k$ nearest neighbors. Suppose we set $k = 8$, and among the eight nearest neighbors, three of them belong to female *Cx. stigmatosoma*, one belongs to female *Ae. aegypti*, and four belong to male *Cx. tarsalis*. We can then calculate the conditional probability using equation (4) as

$$P(\text{X}|\text{female Cx. stigmatosoma}) = \frac{3}{8}$$

$$P(X|\text{female } Ae.aegypti) = \frac{1}{8}$$

$$P(X|\text{male } Cx.tarsalis) = \frac{4}{8}$$

These conditional probabilities are then multiplied by the class *prior* probability to calculate the *posterior* probability, and the observed insect is predicted to be the class which has the highest *posterior* probability. As such, we are able to estimate the class-conditioned probability for features in any format, including the feature of distance returned from an opaque similarity function, and thus generalize the Bayesian classifier to subsume some of the advantages of the nearest neighbor classifier.

Table 1 outlines the Bayesian classification algorithm using the *nearest neighbor distance* feature. The algorithm begins in Lines 1-3 by estimating the *prior* probability for each class. This is done by counting the number of occurrences of each class in the training data set. It then estimates the conditional probability for each unknown data using the kNN approach outlined above. Specifically, given an unknown insect sound, the algorithm first searches the entire training data to find the top *k* nearest neighbors using some distance measure (Lines 5-9); it then counts for each class the number of neighbors which belong to that class and calculates the class-conditioned probability using equation (1). With the prior probability and the class-conditioned probability known for each class, the algorithm calculates the *posterior* probability for each class (Lines 13, 15-18) and predicts the unknown data to belong to the class that has the highest posterior probability (Line 19).

**Table 1: The Bayesian Classification Algorithm Using a High-dimensional Feature**

**Notation**
k: the number of nearest neighbors in kNN approach
disfunc: a distance function to calculate the distance between two data
C: a set of classes
TRAIN: the training dataset
$T_{C_i}$: number of training data that belong to class $C_i$

| | |
|---|---|
| 1 | **for** i = 1 : \|C\| |
| 2 |     P($C_i$) = $T_{C_i}$ / \|TRAIN\|; //estimate *prior* probability |
| 3 | **end** |
| 4 | **for** each unknown data $F_1$ |
| 5 |     **for** j =1: \|TRAIN\| |

```
        d(j) = disfunc(F_1, TRAINj);     //the distance of F_1 to each training data
    end
    [d, sort_index] = sort(d, 'ascend');  //sort the distance in ascending order
    top_k = sort_index(1 to k);           // find the top-k nearest neighbors
    for  i = 1 : |C|
        k_Ci = number of data in top_k that are labeled as class C_i;
        P(F_1|C_i)  =  k_Ci / k;  // calculate the conditional probability with kNN approach
        P(C_i|F_1)  =  P(C_i) P(F_1|C_i);     // calculate posterior probability
    end
    normalize_factor = sum_{i=1}^{i=|C|} P(C_i|F_1)  // normalize the posterior probability
    for i = 1 : |C|
        P(C_i|F_1) = P(C_i|F_1)/normalize_factor;
    end
    Ĉ  = argmax_{C_i∈C} P(C_i|F_1)       // assign the unknown data F_1 to the class Ĉ
end
```

The algorithm outlined in Table 1 requires two inputs, including the parameter *k*. The goal is to choose a value of *k* that minimizes the probability estimation error. One way to do this is to use *validation* (Kohavi 1995). The idea is to keep part of the training data apart as validation data, and evaluate different values of *k* based on the estimation accuracy on the validation data. The value of *k* which achieves the best estimation accuracy is chosen and used in classification. This leaves only the question of which *distance measure* to use, that is, how to decide the distance between any two insect sounds. To find a good distance measure for the flying sounds, we turned to *crowdsourcing* by organizing a contest from July to November in 2012 (Chen et al. 2002) that asked participants to create the best distance measure for the insect sounds. More than fifteen teams worldwide participated in the contest and we received more than eighty submissions (a team is allowed to have multiple submissions and we evaluated each submission, but for the final score, only one submission was scored for a team). The result of the contest suggested that the best distance measure is a simple algorithm which computes the Euclidean distance between the frequency spectrums of the insect sounds. Building on our crowdsourcing efforts, we found that if we truncated the frequency spectrums to exclude data outside the range of possible wingbeat frequencies (cf. Table 2), we could further improve accuracy. Note that some of our crowdsourcing participants had somewhat similar, but less explicit, ideas.

Our distance measure is further explained in Table 2. Given two flying sounds, we first transform each sound into frequency spectrums using DFT (Lines 1-2). The spectrums are then truncated to include only those corresponding to the frequency range from 100 to 2,000 (Lines 3-4); the frequency range is thus chosen, because according to entomological advice[4], all other frequencies are unlikely to be the result of insect activity, and probably reflect noise in the sensor. We then compute the Euclidean distance between the two truncated spectrums (Line 5) and return it as the distance between the two flying sounds.

**Table 2: Our Distance Measure for two Insect Flight Sounds**

**Notation:**
$S_1$,$S_2$: two sound sequences
dis: the distance between the two sounds

| | |
|---|---|
| | **function dis = disfunc($S_1$,$S_2$)** |
| 1 | spectrum1 = DFT(S1); |
| 2 | spectrum2 = DFT(S2); |
| 3 | truncateSpectrum1 = spectrum1(frequency range= [100, 2000]); |
| 4 | truncateSpectrum2 = spectrum2(frequency range= [100, 2000]); |
| 5 | $\text{dis} = \sqrt{\sum(\text{truncateSpectrum}_1 - \text{truncateSpectrum}_2)^2}$ |

Our flying-sounds-based insect classification algorithm is obtained by ‘plugging’ the distance measure explained in Table 2 into the Bayesian classification framework outlined in Table 1. To demonstrate the effectiveness of the algorithm, we considered the data that was used to generate the plot in Figure 1. These data were randomly sampled from a dataset with over 100,000 sounds generated by our sensor. We sampled in total 3,000 flying sounds, 1,000 sounds for each species, so the *prior* probability for each class is one-third. Using our insect classification algorithm with *k* set to eight, which was selected based on the validation result, we achieved an error rate of 12.43% using leave-one-out. We then compared our algorithm to the *optimal* result possible using only the wingbeat frequency, which is the most commonly used approach in previous research efforts. The optimal Bayes error-rate to classify the insects using wingbeat frequency is 18.13%, which is the lower bound for any algorithm that uses just that feature. This means that using the truncated frequency spectrum is able to reduce the error rate by almost a third. To the best of our knowledge, this is the first explicit demonstration that there is exploitable information in the flight sounds beyond the wingbeat frequency.

It is important to note that we do not claim that the distance measure we used in this work is optimal. There may be better measures, which could be discovered by additional research or by revisiting crowdsourcing, etc. Moreover, it is possible that there may be better distance measures if we are confining our attention to just Culicidae or just Tipulidae, etc. However, if and when a better distance measure is found, we can simply ‘plug’ the distance measure in the Bayesian classification framework to get a better classification performance.

[4] Many large insects, i.e. most members of Odonata and/or Lepidoptera, have wingbeat frequencies that are significantly slower than 100 Hz; our choice of truncation level reflects our special interest in Culicidae.

## Additional Feature: Circadian Rhythm of Flight Activity

In addition to the insect flight sounds, there are other features that can be used to reduce the error rate. The features can be very cheap to obtain, as simple as noting the *time-of-intercept*, yet the improvement can be significant.

It has long been noted that different insects often have different circadian flight activity patterns (Taylor 1969), and thus the time when a flying sound is intercepted can be used to help classify insects. For example, house flies (*Musca domestica*) are more active in the daytime than at night, whereas *Cx. tarsalis* are more active at dawn and dusk. If an unknown insect sound is captured at noon, it is more probable to be produced by a house fly than by *Cx. tarsalis* based on this *time-of-intercept* information.

Given an additional feature, we must consider how to incorporate it into the classification algorithm. One of the advantages of using the Bayesian classifier is that it offers a principled way to gracefully combine multiple features; thus, the solution is straightforward here. For simplicity, we temporarily assume the two features, the *insect-sound* and the *time-of-intercept*, are conditionally independent, i.e., they are independent given the class (we will revisit the reasonableness of this assumption later). With such an independence of feature assumption, the Bayesian classifier is called Naïve Bayesian classifier and is illustrated in Figure 6. The two arrows in the graph encode the fact that the probability of an unknown data to be in a class depends on the features $F_1$ and $F_2$, whereas the lack of arrows between $F_1$ and $F_2$ means the two features are independent.

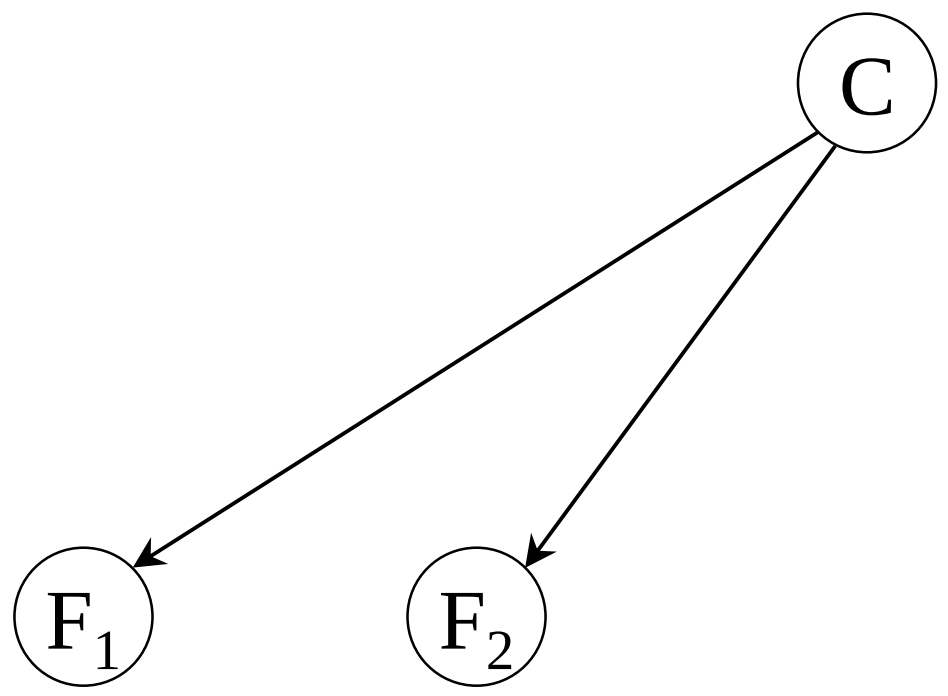

**Figure 6: A Bayesian network that uses two independent features for classification**

An observed object $X$ now should include two values, $f_1$ and $f_2$, and the posterior probability of $X$ belonging to a class $C_i$ is calculated as $\mathrm{P}(C_i|X) = \mathrm{P}(C_i|\mathrm{F}_1 = f_1, \mathrm{F}_2 = f_2)$. With the independence assumption, this probability is proportional to:

$$\mathrm{P}(C_i|\mathrm{F}_1 = f_1, \mathrm{F}_2 = f_2) \propto \mathrm{P}(C_i)\mathrm{P}(\mathrm{F}_1 = f_1|C_i)\mathrm{P}(\mathrm{F}_2 = f_2|C_i) \quad (5)$$

Where $\mathrm{P}\left(\mathrm{F_j} = f_j \middle| \mathrm{C_i}\right)$ is the probability of observing the feature-value pair $\mathrm{F_j} = f_j$ in class $C_i$. For concreteness, the feature $\mathrm{F}_1$ in our algorithm is the insect sound, and $\mathrm{F}_2$ is the time when the sound was produced.

In the previous section, we have shown how to calculate the class-conditioned probability of the *insect-sound*, $P(\mathrm{F}_1 = f_1|C_i)$, using the kNN estimation method and the *prior* probability $P(C_i)$. To incorporate our additional feature, we only need to calculate the class-conditioned probability $P(\mathrm{F}_2 = f_2|C_i)$. Note that the *time-of-intercept* is a scalar. As we discussed in the section above, learning the class-conditioned distributions of a one-dimensional feature can be easily done by constructing a histogram.

This histogram of the *time-of-intercept* for a species is simply the insect's flight activity circadian rhythm. In the literature, there have been many attempts to quantify these patterns for various insects. However, due to the difficulty in obtaining such data, most attempts were made by counting 1 if there is activity observed in a time period (such as a ten-minute window) and 0 otherwise (Taylor and Jones 1969; Rowland and Lindsay 1986). Without distinguishing the number of observations in each period, the resulting patterns have a course granularity. In contrast, by using our sensors we have been able to collect on the order of *hundreds of thousands* of observations per species, and count the *exact* number of observations at sub second granularity, producing what we believe are the “densest” circadian rhythms ever recorded. Figure 7 shows the flight activity circadian rhythms of *Cx. stigmatosoma* (female), *Cx. tarsalis* (male), and *Ae. agypti* (female). Those circadian rhythms were learned based on observations collected over one month. The results are consistent with the report in (Mian et al. 1990; Taylor and Jones 1969), but with a much finer temporal resolution (down to minutes). Note that although all three species are most active at dawn and dusk, *Ae. aegypti* females are significantly more active during daylight hours.

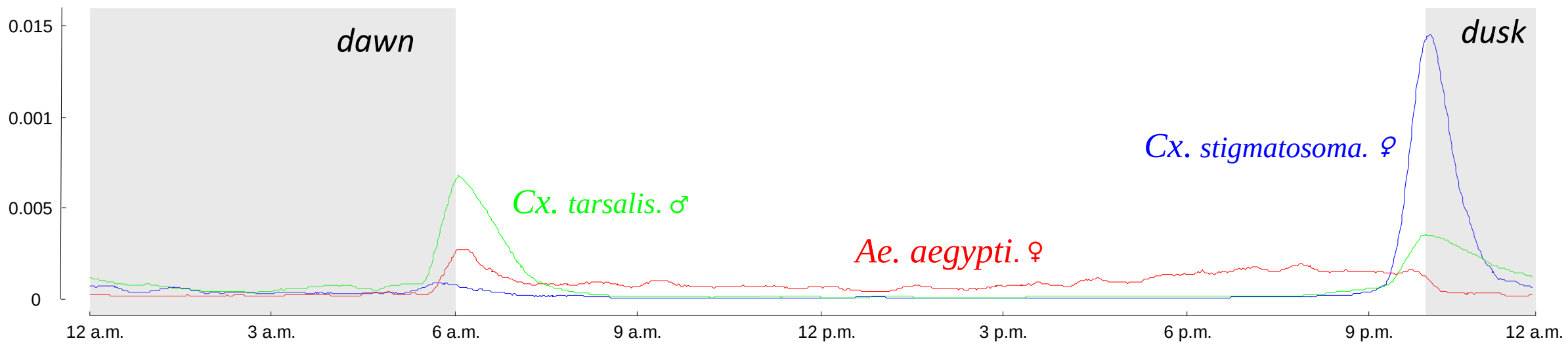

**Figure 7: The flight activity circadian rhythms of *Cx. stigmatosoma* (female), *Cx. tarsalis* (male), and *Ae. Aegypti* (female), learned based on observations generated by our sensor that were collected over one month**

For the insects discussed in this work, we constructed circadian rhythms based on many hundreds of thousands of individual observations. However, it is obvious that as our sensors become more broadly used by the community, we cannot always expect to have such fine-grained data. For example, there are approximately 3,500 mosquito species worldwide; it is unlikely that high quality circadian rhythms for all of them will be collected in the coming years. However, the absence of “gold standard” circadian rhythms should not hinder us from using this useful additional feature. Instead, we may consider using approximate rhythms.

One such idea is to use the circadian rhythms of the most closely related insect species, taxonomically, for which we *do* have data. For example, suppose we do not have the circadian rhythm for *Cx. stigmatosoma*, we can use the rhythm of *Cx. tarsalis* as an approximation if the latter is available.

In the cases where we do not have the circadian rhythms for any taxonomically-related insects, we can construct approximate rhythms simply based on *text* descriptions in the entomological literature. Some frequently encountered descriptions of periods when insects are active include *diurnal*, *nocturnal*, *crepuscular*, etc. Offline, we can build a dictionary of templates based on these descriptions. This process of converting text to probabilities is of course subjective; however, as we will show below, it does lead to improvements over using no time-of-day information. Our simple first attempt at this work is by quantifying different levels of activities with numbers from 1 to 3, representing, *low*, *medium*, and *high* activity[5]. For example, if an insect is described as diurnal and most active at dawn and dusk, we can use these three degrees to quantify the activities: highest degree at dawn and dusk, second in the daytime, and low activity during the night. The resulting template is shown in Figure 8.IV. Note that a circadian rhythm is a probability distribution that tells how likely we are to capture a certain insect species' flights at a certain time, and thus each template is normalized such that the area under the template sums to one. In Figure 8.II, we show an approximate circadian rhythm for *Cx. stigmatosoma* that we constructed this way. We spend two minutes searching the web for an academic paper that describes *Cx. stigmatosoma* flight activity, discovering that according to (Mian et al. 1990), *Cx. stigmatosoma* is active at dawn and dusk (*crepuscular*).

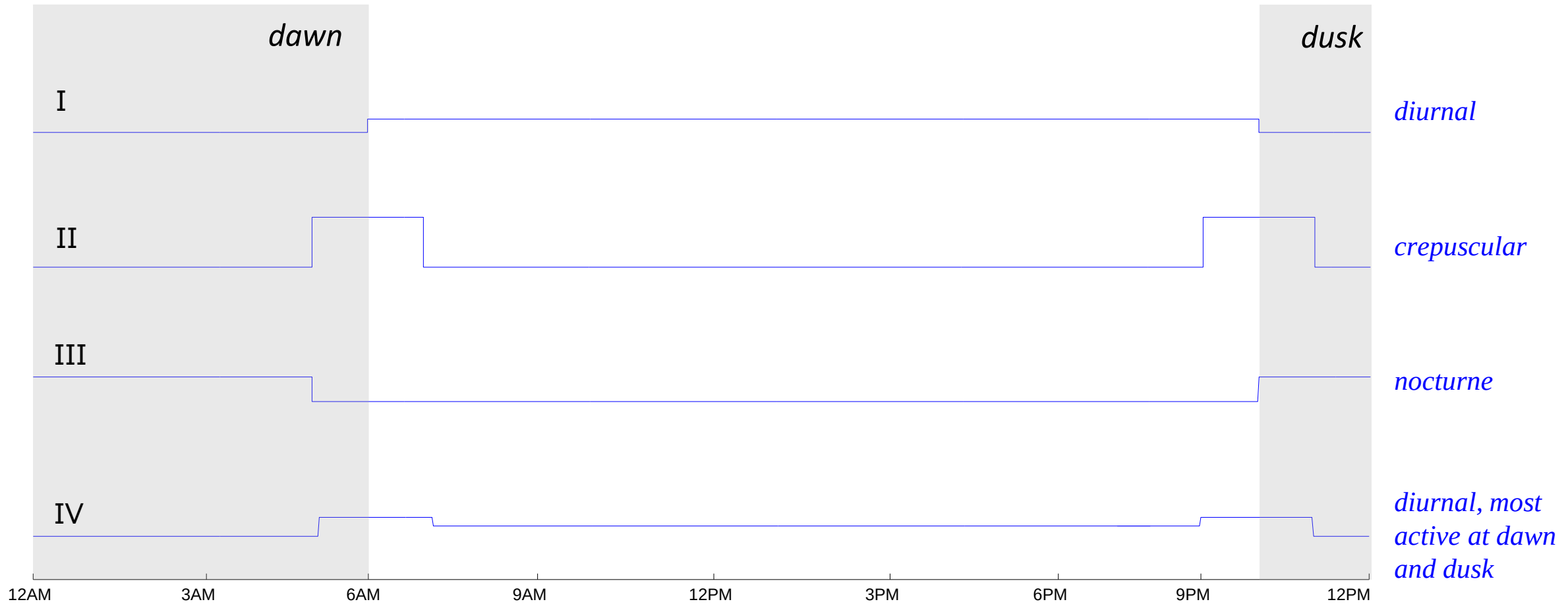

**Figure 8: Examples of approximation templates for insects' flight activity circadian rhythm. No markings are shown on the Y-axis; all templates are normalized such that the area under the curve is one, and the smallest value is an epsilon greater than zero**

In the worst case, if we cannot glean any knowledge about the insect's circadian rhythm (e.g. a newly-discovered species), we can simply use a constant line as an approximation. The constant line encodes no information about the activity hours of that insect, but it enables the incorporation of the more familiar insects' circadian rhythms into the classifier to improve performance. In the pathological case where all the circadian rhythms are approximated using a constant line, the classifier degenerates to a Bayesian classifier that does not use this additional feature.

[5] The lowest level of flight activity we represent is *low*, but not *zero*. In a Bayesian classifier, we never want to assign zero probabilities to an event, unless we are sure it is logically impossible to occur. In practice, a technique called Laplacian correction is typically used to prevent any probability estimate from being exactly zero.

Given the above, we can almost always incorporate some of the circadian rhythm information into our classifier. Given a *time-of-intercept* observation, we can calculate the class-conditioned probability of observing the activity from a species simply by looking up the flight activity circadian rhythms of that species. For example, suppose an insect sound was detected at 6:00am, the probability of observing this activity from a *Cx. tarsalis* male is three times the probability of observing it from an *Ae. aegypti* female, according to the circadian rhythms shown in Figure 7.

For concreteness, the insect classification algorithm that uses two features is outlined in Table 3. It is similar to the algorithm outlined in Table 1 that uses a single feature. Only five modifications are made.

**Table 3: The Insect Classification Algorithm Using two Features**

This algorithm is similar to the one outlined in Table 1. Only three modifications are needed, which are listed below. The modifications are highlighted in blue/bold.

| | | |
|---|---|---|
| 1 | Line 6: | for each unknown data $[\mathbf{F_1}, \boldsymbol{F_2}]$ |
| 2 | Line 13: | $P(C_i\|\mathbf{F_1}, \mathbf{F_2}) = P(C_i)\, P(F_1\|C_i)\, \mathbf{P(F_2\|\boldsymbol{C_i})}$; // calculate the *posterior* probability |
| 3 | Line 15: | normalize_factor = $\sum_{i=1}^{i=\|C\|} \mathbf{P(\boldsymbol{C_i}\|F_1, F_2)}$ // normalize the posterior probability |
| 4 | Line 17: | $\mathbf{P(\boldsymbol{C_i}\|F_1, F_2)} = \mathbf{P(\boldsymbol{C_i}\|F_1, F_2)}$/normalize_factor; |
| 5 | Line 19: | $\hat{C} = \underset{C_i \in C}{\operatorname{argmax}}\, \mathbf{P(C_i\|F_1, F_2)}$ // assign the unknown data X to the class $\hat{C}$ |

To demonstrate the benefit of incorporating the additional feature in classification, we again revisit the toy example in Figure 1. With the *time-of-intercept* feature incorporated and the accurate flight activity circadian rhythms learned using our sensor data, we achieve a classification accuracy of 95.23%. Recall that the classification accuracy using *just* the *insect-sound* is 87.57% (cf. the paragraph right below Table 2). Simply by incorporating this cheap-to-obtain feature, we reduce the classification error rate by about two-thirds, from 12.43% to only 4.77%.

To test the effect of using proxies of the learned flight activity circadian rhythm, we imagine that we do not have the flight activity circadian rhythm for *Cx. stigmatosoma* female, and that we must use one of the approximate rhythms discussed above. The results are shown in Table 4. As we can see, even with a constant line approximation, the classification accuracy is 88.73%, slightly better than not using the *time-of-intercept* feature. This is because, although the algorithm has no knowledge about the circadian rhythm for *Cx. stigmatosoma* females, it does have knowledge of the other two species' circadian rhythms. With the approximation created based on the text description, we achieve an accuracy of 90.80%, which is better than using the constant line. This is as we hoped, as the approximate rhythm carries some useful information about the insects' activity pattern, even though it is at a very coarse granularity. An even better classification accuracy of 93.87% is achieved by using the circadian rhythm of *Cx. tarsalis* males as the approximation. As can be seen from Figure 7, the circadian rhythm of *Cx. tarsalis* males is quite close to that of *Cx. stigmatosoma* females.

**Table 4: Classification Performance using Different Approximations for *Cx. stigmatosoma* (female) Flight Activity Circadian Rhythm**

| **Flight activity circadian rhythm approximations** | No rhythm used | constant line approximation | description based approximation | using the taxonomically-related insect's rhythm | learned using our sensor data |
|---|---|---|---|---|---|
| **Classification Accuracy** | 87.57% | 88.73% | 90.80% | 93.87% | 95.23% |

Note that the classification accuracy with *any* approximation is worse than that of using the accurate circadian rhythm (95.23% accuracy). This is not surprising, as the more accurate the estimated distribution is, the more accurate the classification will be. This reveals the great utility of our sensor: it allows the inexpensive collection of massive amounts of training data, which can be used to learn accurate distributions.

## A Tentative Additional Feature: Geographic Distribution

In addition to the *time-of-intercept*, we can also use the *location-of-intercept* as an additional feature to reduce classification error rate. The *location-of-intercept* is also very cheap-to-obtain. It is simply the location where the sensor is deployed.

We must preempt a possible confusion on behalf of the reader. One application of our sensors is estimating the relative abundance of various species of insects at some particular location. However, we are suggesting here that we can use estimates of the relative abundance of the species of insects at that location to do this more accurately. This appears to be a “chicken and egg paradox.” However, there *is* no contradiction. The classifier is attempting to optimize the accuracy of its *individual* decisions about each particular insect observed, and knowing, even approximately, the expected prevalence of each species can improve accuracy.

The *location-of-intercept* carries useful information for classification because insects are rarely evenly distributed at any spatial granularity we consider. For example, *Cx. tarsalis* is relatively rare east of the Mississippi River (Reisen 1993), whereas *Aedes albopictus* (the Asian tiger mosquito) has now become established in most states in that area (Novak 1992). If an insect is captured in some state east of the Mississippi River, it is more probable to be an *Ae. albopictus* than a *Cx. tarsalis*. At a finer spatial granularity, we may leverage knowledge such as “since this trap is next to a dairy farm (an animal manure source), we are five times more likely to see a *Sylvicola fenestralis* (Window Gnat), than an *Anopheles punctipennis.*”

A Bayesian classifier that uses three features is illustrated in Figure 9. Here, we again assume that all the three features, *insect-sound*, *time-of-intercept*, and *location-of-intercept*, are independent.

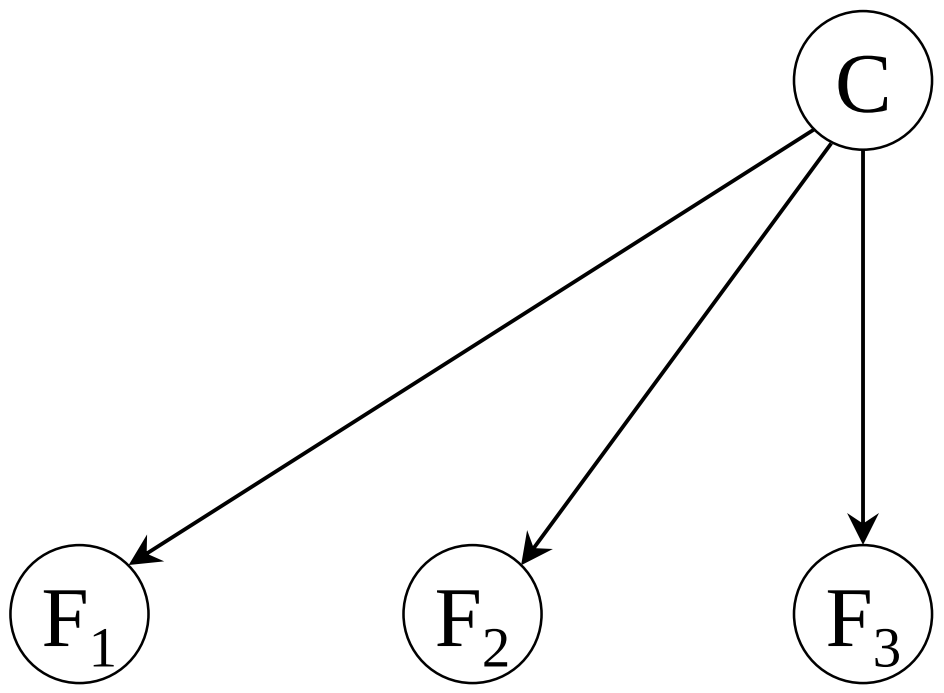

**Figure 9: A Bayesian network that uses three independent features for classification**

Based on Figure 9, the probability of an observed object $X$ belonging to a class $C_i$ is calculated as:

$$\mathrm{P}(C_i|\mathrm{F}_1 = f_1, \mathrm{F}_2 = f_2, \mathrm{F}_3 = f_3) \propto \mathrm{P}(C_i)\mathrm{P}(\mathrm{F}_1 = f_1|C_i)\mathrm{P}(\mathrm{F}_2 = f_2|C_i)\mathrm{P}(\mathrm{F}_3 = f_3|C_i) \quad (6)$$

Where $P(\mathrm{F}_3 = f_3|C_i)$ is the probability of observing an insect from class $C_i$ at location $f_3$. This probability reflects the geographic distribution of the insects. For classification, we do not need the true *absolute* densities of insect prevalence; we just need the *ratio* of densities for the different species at the observation location. This is because with the ratio of $P(\mathrm{F}_3 = f_3|C_i)$, we can calculate the ratio of the posterior probability $\mathrm{P}(C_i|\mathrm{F}_1 = f_1, \mathrm{F}_2 = f_2, \mathrm{F}_3 = f_3)$ of each species, and predict an observation to belong to the species which has the highest posterior probability. In the case where we do need the actual posterior probability values, we can always calculate them from the posterior probability ratio, based on the constraint that the sum of all posterior probabilities over different classes should be one, as shown in Lines 15-18 in Table 1.

To obtain the ratio of $\mathrm{P}(\mathrm{F}_3 = f_3|C_i)$ at a given a location is simple. We can glean this information from the text of relevant journal papers or simply from the experiences of local field technicians. For example, suppose we deploy our insect sensor at a location where we should expect to be twice as likely to encounter *Cx. tarsalis* as *Cx. stigmatosoma*, the ratio of *Cx. tarsalis* to *Cx. stigmatosoma* is 2:1. In the case where we cannot glean any such knowledge about the local insect population, we can temporarily augment our sensor with various insect traps that physically capture insects (i.e. CDC trap for mosquitoes, "yellow sticky cards" traps for sharpshooters, etc.), and use these manually counted number of observations of each species to estimate the ratios.

To demonstrate the utility of incorporating this *location-of-intercept* feature, we did a simple *simulation* experiment. Note that the *insect-sound* and *time-of-intercept* features are *real* data; only the *location-of-intercept* was simulated by assuming there are two species of insects, *Cx. stigmatosoma* (female) and *Ae. aegypti* (female), which are geographically distributed as shown in Figure 10. We further assumed our sensors are deployed at three different locations, $S_1$, $S_2$ and $S_3$, where $S_2$ is about the same distance from both centers, and $S_1$ and $S_3$ are each close to one of the centers. Here, we model the location distributions as Gaussian density "bumps" for simplicity; however, this is not a necessary assumption, but we can use any density distribution.

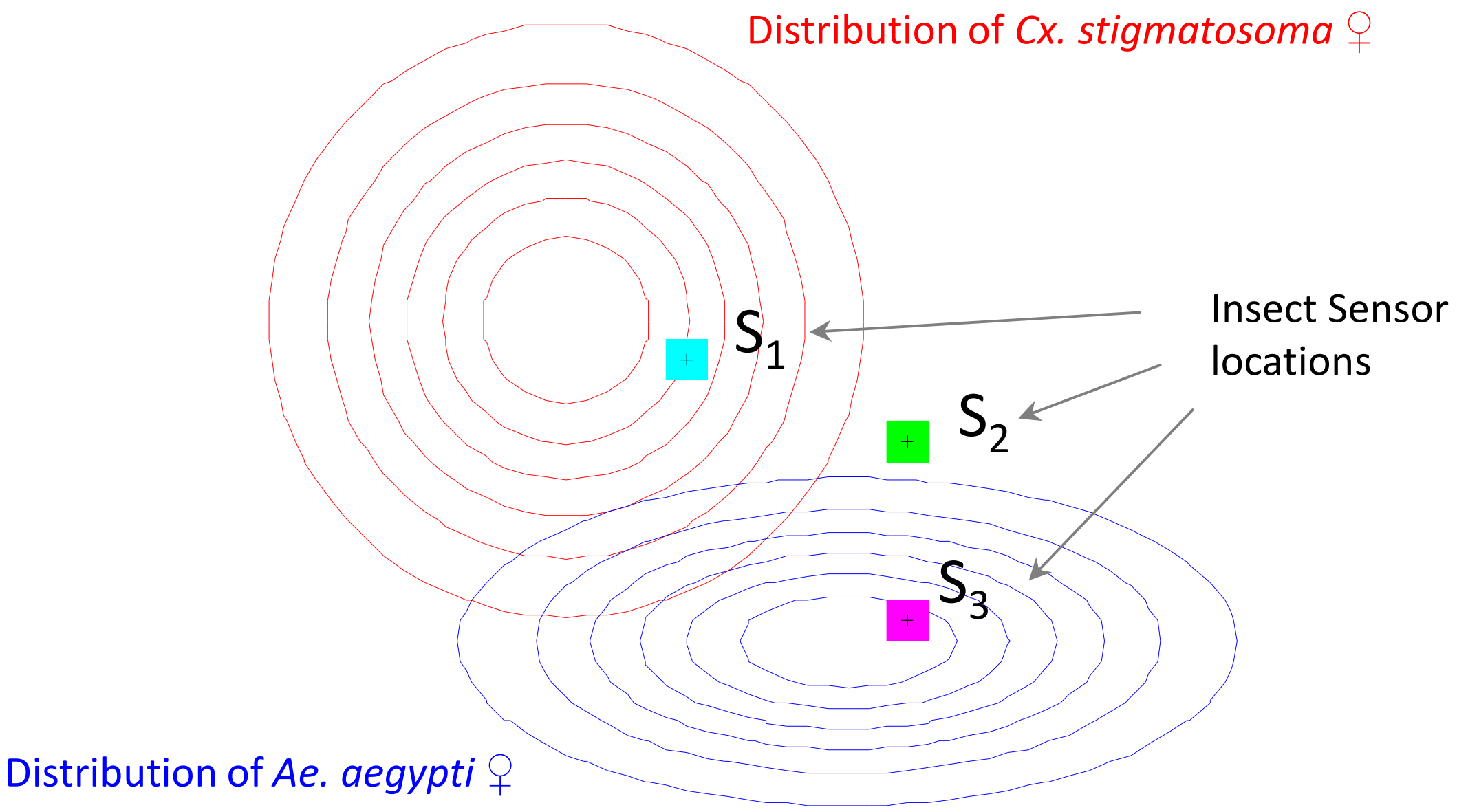

**Figure 10: The assumptions of geographic distributions of each insect species and sensor locations in our simulation to demonstrate the effectiveness of using *location-of-intercept* feature in classification**

To simulate the data captured by the three sensors, we project ten thousand insect exemplars of each species onto the map according to the geographic distribution assumption. We then sample these insects that are within the capture range of the sensors, which is assumed to be a square region centering at the sensor, as shown in Figure 10. Each sampled insect is assumed to fly across our sensor once and have its data captured.

In our experiment, we sampled 277 *Cx. stigmatosoma* females and 3 *Ae. aegypti* females at location $S_1$, 24 and 44 at location $S_2$, and 4 and 544 at location $S_3$. Using just the *frequency spectrum* and the *time-of-intercept* to classify those sampled insects, we achieved an error rate of 5.41%. However, by incorporating the *location-of-intercept*, we reduced the error rate to 2.54%. This is impressive, as the *location-of-intercept* information is very cheap-to-obtain, yet it reduced the error rate by more than half.

## A General Framework for Adding Features

In the previous two sections, we have shown how to extend our insect flight sound classifier to incorporate two additional features. However, there may be dozens of additional features that could help improve the classification performance. These potential features are so domain and application specific that we cannot give any more than a few representative examples here. It has long been known that some species of insects have a preferred height at which they fly. For example, (Njie et al. 2009) noted that *Anopheline* mosquitoes are much likely to be observed than *culicine* flying at a height two meters, which is the approximate height required to enter through the eave of a house. Here, we imagine using a *height-of-intercept* feature. By placing two of our sensors side-by-side at a known distance apart and observing the lag between the sensor observations, we can obtain an approximation of the speed at which the insect is flying (only an approximation, as the insect may fly at an angle to the light beam). This *speed-of-intercept* feature may help to discriminate between speedy members of the genus *Culicoides*, flying at 250cm/s and the relatively sluggish members of the family Culicidae which max out at about 120cm/s (Bidlingmayer et al. 1995).

In this section, we generalize our classifier to a framework that is easily extendable to incorporate arbitrarily many such specialized features.

If we compare Figure 4, Figure 6, and Figure 9, which show the Bayesian networks with an increasing number of features, we can see that adding a feature to the classifier is represented by adding a node to the Bayesian network. The Bayesian network that uses $n$ independent features for classification is shown in Figure 11.

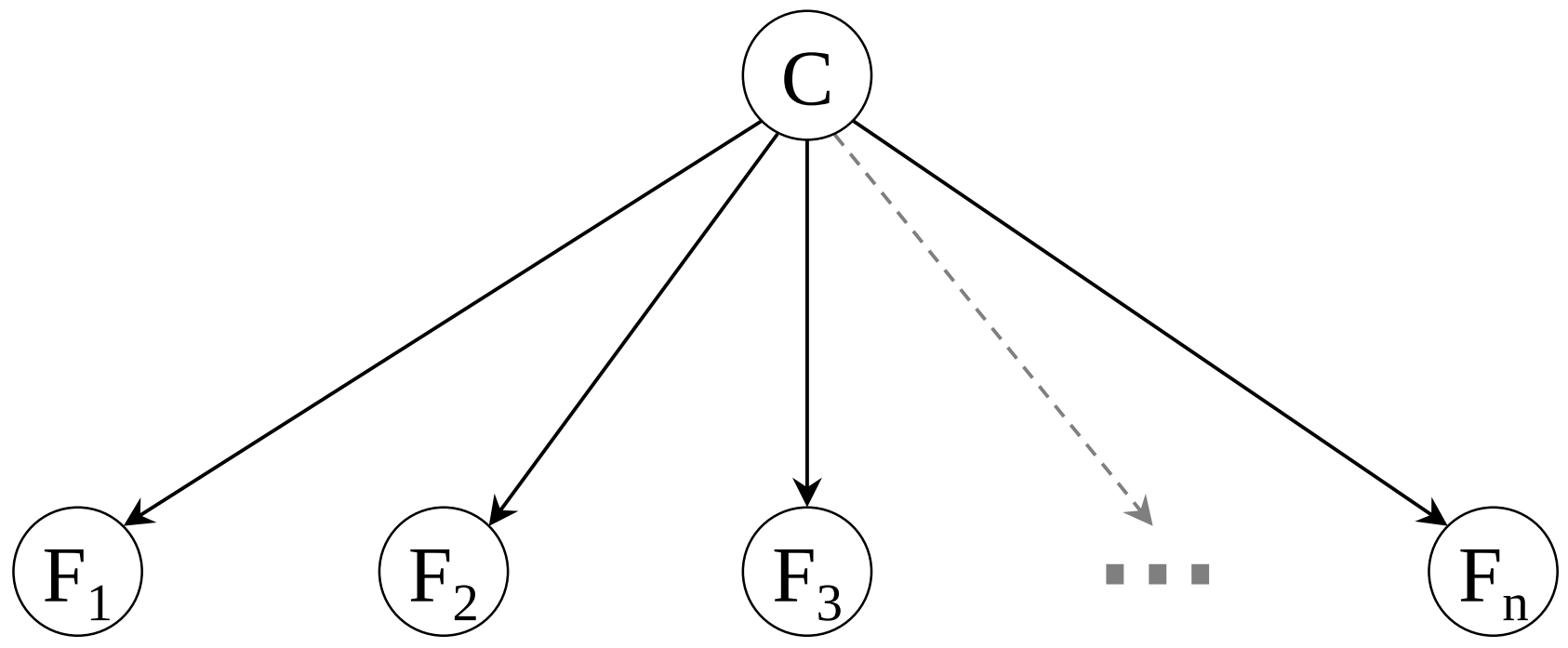

**Figure 11: The general Bayesian network that uses n features for classification, where n is a positive integer**

With $n$ independent features, the posterior probability that an observation belongs to a class $C_i$ is calculated as:

$$P(C_i | F_1 = f_1, F_2 = f_2, \dots, F_n = f_n) \propto P(C_i) \prod_{j=1}^{n} P(F_j = f_j | C_i) \qquad (7)$$

Where $\mathrm{P}(\mathrm{F_j} = f_j|C_i)$ is the probability of observing $f_j$ in class $C_i$.

Note that the posterior probability can be calculated incrementally as the number of features increases. That is, if we have used some features to classify the objects, and later on, we have discovered more useful features and would like to add those new features to the classifier to re-classify the objects, we do not have to re-compute the entire classification from scratch. Instead, we can keep the posterior probability obtained from the previous classification (based on the old features), update each posterior probability by multiplying it with the corresponding class-conditioned probability of the new features, and re-classify the objects using the new posterior probabilities. For example, suppose we first used the *insect-sound* and *time-of-intercept* to classify an observation $X$. The posterior probability of $X$ belonging to *Ae. aegypti* is 0.4 and that of belonging to *Cx. tarsalis* is 0.6. Later on, we find that *location-of-intercept* is useful and would like to incorporate this new feature to re-classify $X$. Suppose the probability of observing *Ae. aegypti* at the location where $X$ was intercepted was 0.5, and the probability of observing a *Cx. tarsalis* at that location was 0.25. In that case, we can update the posterior probability of $X$ belonging to *Ae. agypti* as $0.4 \times 0.5 = 0.2$, and that of belonging to *Cx. tarsalis* as $0.5 \times 0.25 = 0.125$, and (in this case) re-predict $X$ to belong to *Ae. agypti*. The advantage of incremental calculation is that incorporating a new feature is fast; only a simple multiplication is required to calculate a posterior probability.

In our discussions thus far, we have assumed that all the features are independent given the class. In practice, features are seldom independent given the class. However, as shown in (Domingos and Pazzani 1997), even when the independence assumption does not hold, the naïve Bayesian classifier may still be optimal in minimizing the misclassification error. Empirical evidence in recent years also showed that the Naïve Bayesian classifier works quite well even in the domains where clear feature dependences exist. In this work, we will not prove that the three features used are conditionally independent. However, as we shall show below, the independence assumption of the features used in our classifier should be reasonable for the Bayesian classifier to work well.

## Revisiting the Independent Assumption

Recall that the naive Bayesian classifier is optimal only under the assumption that the features are independent (Domingos and Pazzani 1997). The majority of experiments in this work consider two features, *frequency spectrum* (*insect-sound*) and *time-of-intercept*. In order to test if these two features $F_1$ and $F_2$ are conditionally independent, we can check if they satisfy the constraint

$$P(F_1|C_i) \cong P(F_1|C_i, F_2)$$

Concretely, for our task, this constraint is $P(spectrum) \cong P(spectrum|time)$ given a certain insect species. Or equivalently, the properties of *frequency spectrum* of a given species must be the same at any timestamp. If the spectrum was a *scalar* value (such as *mass* or *length*), we could use a standard test such as a Kolmogorov–Smirnov test to see if the properties observed at two different time windows are from the same distribution. However, the spectrums are vectors, and this complicates this issue greatly. Thus, to see if this constraint is satisfied, we did the following experiment which indirectly but forcefully tests the constraint. We sampled 1,000 insect sounds captured at dawn (between 5:00am-8:00am) and 1,000 sounds captured at dusk (between 8pm-11pm). All the sounds were generated by *Ae. aegypti* females. We then classified the sounds captured in different time periods using the *frequency spectrum*. Our hypothesis is that, if the distribution of *frequency spectrum* of a same species is the same at dawn and at dusk, then it should be impossible to distinguish between sounds captured in the two different periods, and thus the classification error rate would be around 50%. In our experiment, the sounds were sampled from a pool of 10,000 objects, and we averaged over ten samplings with replacement. The average classification error rate was 50.14%, which suggests that there is no perceptible difference in the *frequency spectrum* of the insect sounds captured at dawn or at dusk.

Note that this experiment was conducted for insects observed under constant temperature and humidity in an insectary. It may not generalize to insects observed in the field. However, this experiment increases our confidence that the conditional independence assumption of the two features is at least reasonable. Nevertheless, it is clear that in our general framework, it is possible that users may wish to use features that clearly violate this assumption. For example, if the sensor was augmented to obtain *insect mass* (a generally useful feature), it is clear from basic principles of allometric scaling that the frequency spectrum feature would *not* be independent (Deakin 2010). The good news is that as shown in Figure 12, the Bayesian network can be generalized to encode the dependencies among the features. In the cases where there is clear dependence between some features, we can consider adding an arrow between the dependent features to represent this dependence. For example, suppose there is dependence between features $F_2$ and $F_3$, we can add an arrow between them, as shown by the red arrow in Figure 12. The direction of the arrow represents *causality*. For example, an insect's larger *mass* causes it to have a slower wingbeat. The only drawback to this augmented Bayesian classifier (Keogh and Pazzani 1999) is that more training data is required to learn the classification model if there are feature dependences, as more distribution parameters need to be estimated (e.g., the covariance matrix is required instead of just the standard deviation) .

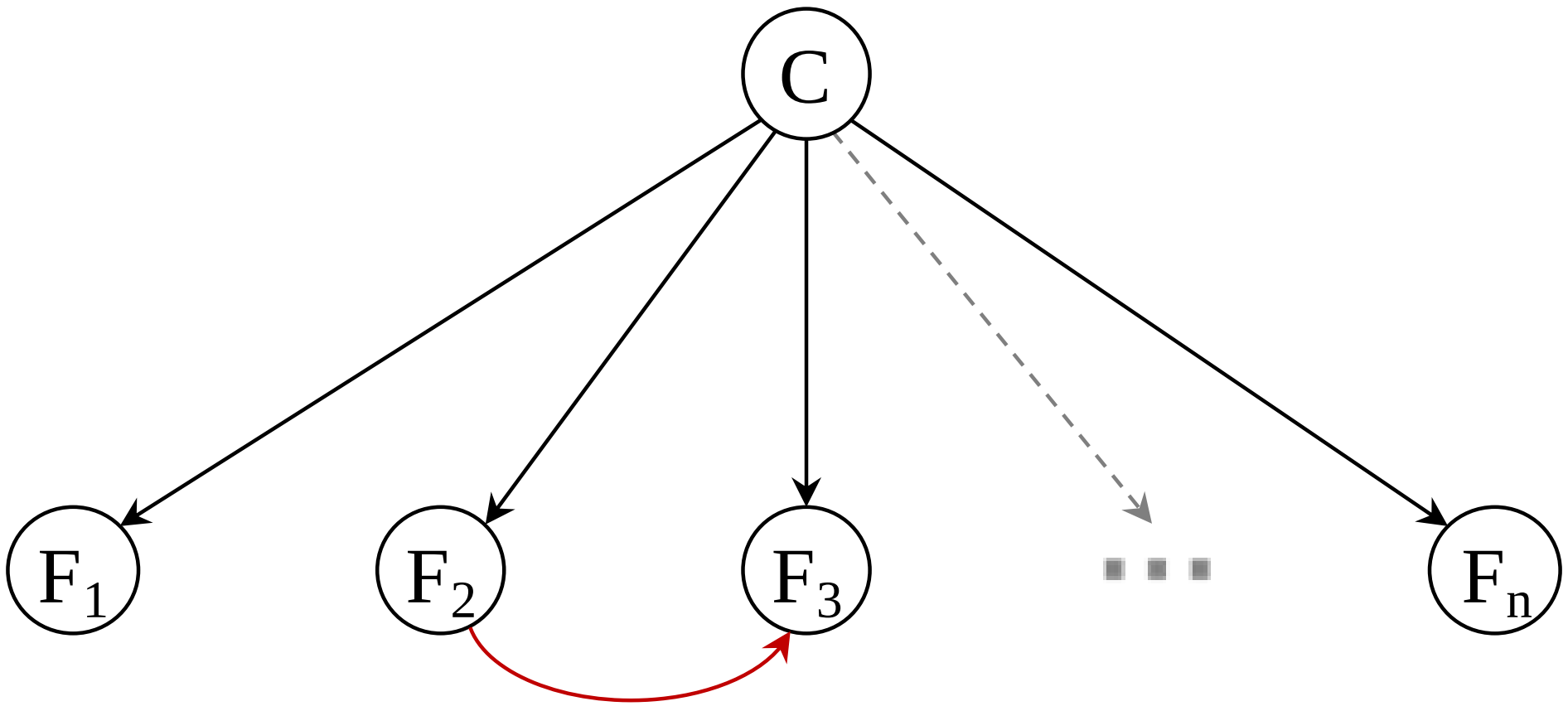

**Figure 12: The Bayesian network that uses n features for classification, with feature $F_2$ and $F_3$ being conditionally dependent**

## A Case Study: Sexing Mosquitoes

Sexing mosquitoes is required in some entomological applications. For example, Sterile Insect Technique, a method which eliminates large populations of breeding insects by releasing only sterile males into the wild, has to separate the male mosquitoes from the females before being released (Papathanos et al. 2009). Here, we conducted an experiment to see how well it is possible to distinguish female and male mosquitoes from a single species using our proposed classifier.

In this experiment, we would like to distinguish male *Ae. aegypti* mosquitoes from females. The only feature used in this experiment is the *frequency spectrum*. We did not use the *time-of-intercept*, as there is no obvious difference between the flight activity circadian rhythms of the males and the females that belong to a same species (A recent paper offers evidence of minor, but measurable differences for the related species *Anopheles gambiae* (Rund et al. 2012); however, we ignore this possibility here for simplicity). The data used were randomly sampled from a pool of over 20,000 exemplars. We varied the number of exemplars from each sex from 100 to 1,000 and averaged over 100 runs, each time using random sampling with replacement. The average classification performance using leave-one-out cross validation is shown in Figure 13.

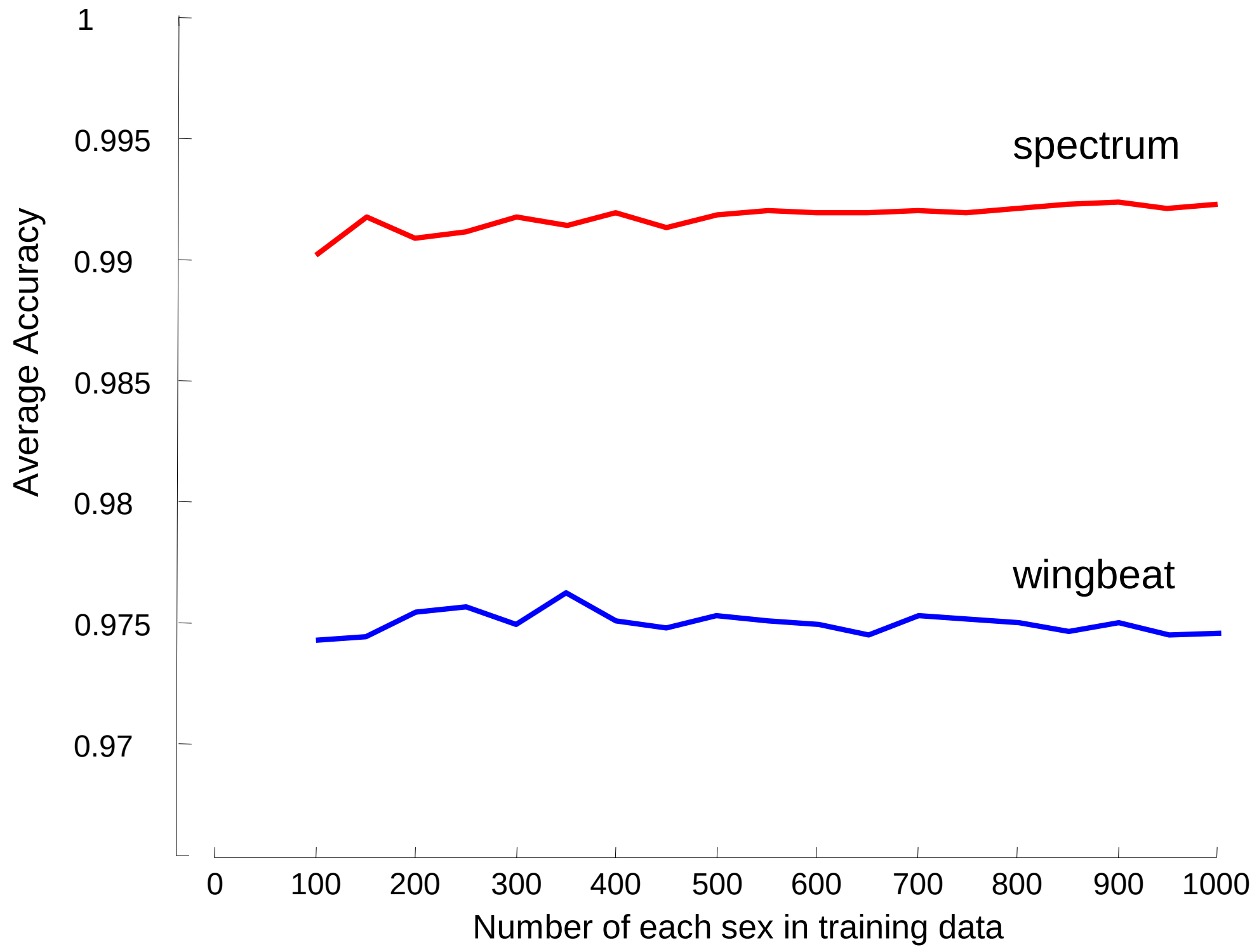

**Figure 13: The classification accuracy of sex discrimination of *Ae. agypti* mosquitoes with different numbers of training data using our proposed classifier and the wingbeat-frequency-only classifier.**

We can see that our classifier is quite accurate in sex separation. With 1,000 training data for each sex, we achieved a classification accuracy of 99.22% using *just* the truncated frequency spectrum. That is, if our classifier is used to separate 1,000 mosquitoes, we will make about eight misclassifications. Note that, as the amount of training data increases, the classification accuracy increases. This is an additional confirmation of the claim that more data improves classification (Halevy et al. 2009).

We compared our classifier to the classifier using just the wingbeat frequency. As shown in Figure 13, our classifier consistently outperforms the wingbeat frequency classifier across the entire range of the number of training data. The classification accuracy using the wingbeat classifier was 97.47% if there are 1,000 training data for each sex. Recall that the accuracy using our proposed classifier was 99.22%. By using the frequency spectrum instead of the wingbeat frequency, we reduced the error rate by more than two-thirds, from 2.53% to 0.78%. It is important to recall that in this comparison, the data and the basic classifier were *identical*; thus, all the improvement can be attributed to the additional information available in the frequency spectrum beyond just the wingbeat frequency. This offers additional evidence for our claim that wingbeat frequency by itself is insufficient for accurate classification.

In this experiment, we assume the cost of female misclassification (misclassifying a female as a male) is the same as the cost of male misclassification (misclassifying a male as a female). The confusion matrix of classifying 2,000 mosquitoes (equal size for each sex) with the same cost assumption from one experiment is shown in Table 5. I.

**Table 5: (I) The confusion matrix for sex discrimination of *Ae. aegypti* mosquitoes with the decision threshold for female being 0.5 (i.e., same cost assumption). (II) The confusion matrix of sexing the same mosquitoes with the decision threshold for female being 0.1**

| I (Balanced cost) | | Predicted class: female | Predicted class: male |
|---|---|---|---|
| Actual class | female | 993 | 7 |
| | male | 5 | 995 |

| II (Asymmetric cost) | | Predicted class: female | Predicted class: male |
|---|---|---|---|
| Actual class | female | 1,000 | 0 |
| | male | 22 | 978 |

However, there are cases in which the misclassification costs are asymmetric. For example, when the Sterile Insect Technique is applied to mosquito control, failing to release an occasional male mosquito because we mistakenly thought it was a female does not matter too much. In contrast, releasing a female into the wild is a more serious mistake, as it is only the females that pose a threat to human health. In the cases where we have to deal with asymmetric misclassification costs, we can change the decision boundary of our classifier to lower the number of high-cost misclassifications in a principled manner. Of course, there is *no free lunch*, and a reduction in the number of high-cost misclassifications will be accompanied by an increase in the number of low-cost misclassifications.

In the previous experiment, with equal misclassification costs, an unknown insect is predicted to belong to the class that has the higher posterior probability. This is the equivalent of saying the threshold to predict an unknown insect as female is 0.5. That is, only when the posterior probability of belonging to the class of females is larger than 0.5 will an unknown insect be predicted as a female. Equivalently, we can replace Line 19 in Table 1 with the code in Table 6 by setting the threshold to 0.5.

**Table 6: The decision making policy for the sex separation experiment**

```
if ( P(female|X) ≥ threshold )
    X is a female
else
    X is a male
end
```

We can change the threshold to minimize the *total* cost when the costs of different misclassifications are different. In Sterile Insect Technique, the goal is to reduce the number of female misclassifications. This can be achieved by lowering the threshold required to predict an exemplar to be female. For example, we can set the threshold to be 0.1, so that if the probability of an unknown exemplar belonging to a female is no less than this value, it is predicted as a female. While changing the threshold may result in a lower overall accuracy, as more males will be misclassified as females, it reduces the number of females that are misclassified as male. By examining the experiment summarized in Table 5. I, we can predict that by setting the threshold to be 0.1, we reduce the female misclassification rate to 0.075%, with the male misclassification rate rising to 0.69%. We chose this threshold value because it gives us an approximately one in a thousand chance of releasing a female. However, any domain specific threshold value can be used; the practitioner simply needs to state her preference in one of two intuitive and equivalent ways: "*What is the threshold that gives me a one in* (some value) *chance of misclassifying a female as a male*" or "*For my problem, misclassifying a male as a female is* (some value) *times worse than the other type of mistake, what should the threshold be?*" (Elkan 2001).

We applied our 0.1 threshold to the data which was used to produce the confusion matrix shown in Table 5.I and obtained the confusion matrix shown in Table 5.II. As we can see, of 2,000 insects in this experiment, twenty-two males, and *zero* females where misclassified, numbers in close agreement to theory.

## Experiment: Insect Classification with Increasing Number of Species

When discussing our sensor/algorithm, we are invariably asked, "*How accurate is it*?" The answer to this depends on the insects to be classified. For example, if the classifier is used to distinguish *Cx. stigmatosoma* (female) from *Cx. tarsalis* (male), it can achieve near perfect accuracy as the two classes are radically different in their wingbeat sounds; whereas when it is used to separate *Cx. stigmatosoma* (female) from *Ae. aegypti* (female), the classification accuracy will be much lower, given that the two species have quite similar sounds, as hinted at in Figure 1. Therefore, a single absolute value for classification accuracy will not give the reader a good intuition about the performance of our system. Instead, in this section, rather than reporting our classifier's accuracy on a fixed set of insects, we applied our classifier to datasets with an incrementally increasing number of species and therefore increasing classification difficulty.

We began by classifying just two species of insects; then at each step, we added one more species (or a single *sex* of a sexually dimorphic species) and used our classifier to classify the increased number of species. We considered a total of ten classes of insects (different sexes from the same species counting as different classes), 5,000 exemplars in each class. Our classifier used both *insect-sound* (frequency spectrum) and *time-of-intercept* for classification. The classification accuracy measured at each step and the relevant class added is shown in Table 7. Note that the classification accuracy at each step is the accuracy of classifying all the species that come *at* and *before* that step. For example, the classification accuracy at the last step is the accuracy of classifying *all* ten classes of insects.

**Table 7: Classification accuracy with increasing number of classes**

| Step | Species Added | Classification Accuracy | | Step | Species Added | Classification Accuracy |
|---|---|---|---|---|---|---|
| 1 | *Ae. aegypti* ♂ | N/A | | 6 | *Cx. quinquefasciatus* ♂ | 92.69% |
| 2 | *Musca domestica* | 98.99% | | 7 | *Cx. stigmatosoma* ♀ | 89.66% |
| 3 | *Ae. aegypti* ♀ | 98.27% | | 8 | *Cx. tarsalis* ♂ | 83.54% |
| 4 | *Cx. stigmatosoma* ♂ | 97.31% | | 9 | *Cx. quinquefasciatus*♀ | 81.04% |
| 5 | *Cx. tarsalis* ♀ | 96.10% | | 10 | *Drosophila simulans* | 79.44% |

As we can see, our classifier achieves more than 96% accuracy when classifying no more than five species of insects, significantly higher than the default rate of 20% accuracy. Even when the number of classes considered increases to ten, the classification accuracy is never lower than 79%, again significantly higher than the default rate of 10%. Note that the ten classes are not easy to separate, even by human inspection. Among the ten species, eight of them are mosquitoes; six of them are from the same genus.

## The Utility of Automatic Insect Classification

The reader may already appreciate the utility of automatic insect classification. However, for completeness, we give some examples of how the technology may be used.

- Electrical Discharge Insect Control Systems EDICS ("bug zappers") are insect traps that attract and then electrocute mosquitoes. They are very popular with consumers who are presumably gratified when hearing the characteristic buzz sound produced when an insect is electrocuted. While most commercial devices are sold as mosquito deterrents, studies have shown that as little as 0.22% of the insects killed are mosquitoes (Frick and Tallamy 1996). This is not surprising, since the attractant is typically just an ultraviolet light. Augmenting the traps with $CO_2$ or other chemical attractants helps, but still allows the needless electrocution of beneficial insects. ISCA technologies (owned by author A. M-N) is experimenting with building a "smart trap" that classifies insects as they approach the trap, selectively killing the target insects but blowing the non-target insects away with compressed air.
- As noted above, Sterile Insect Technique has been used to reduce the populations of certain target insects, most notably with Screwworm flies (Cochliomyia hominovorax) and the Mediterranean fruit fly (Ceratitis capitata). The basic idea is to release sterile males into the wild to mate with wild females. Because the males are sterile, the females will lay eggs that are either unfertilized, or produce a smaller proportion of fertilized eggs, leading to population declines and eventual eradication in certain areas. (Benedict and Robinson 2003). Note that it is important not to release females, and sexing mosquitoes is notoriously difficult. Researchers at the University of Kentucky are experimenting with our sensors to create insectaries from which only male hatchlings can escape. The idea is to use a modified EDICS or a high powered laser that selectively turns on and off to allow males to pass through, but kills the females.

- Much of the research on insect behavior with regard to color, odor, etc., is done by having human observers count insects as they move in dual choice olfactometer or on landing strips etc. For example, (Cooperband et al. 2013) notes, "*Virgin female wasps were individually released downwind and the color on which they landed was recorded* (by a human observer)." There are several problems with this: human time becomes a bottleneck in research; human error is a possibility; and for some host seeking insects, the presence of a human nearby may affect the outcome of the experiment (unless costly isolation techniques/equipment is used). We envision our sensor can be used to accelerate such research by making it significantly cheaper to conduct these types of experiments. Moreover, the unique abilities of our system will allow researchers to conduct experiments that are currently impossible. For example, a recent paper (Rund et al. 2012) attempted to see if there are sex-specific differences in the daily flight activity patterns of *Anopheles gambiae* mosquitoes. To do this, the authors placed individual sexed mosquitoes in small glass tubes to record their behavior. However, it is possible that both the small size of the glass tubes and the fact that the insects were in isolation affected the result. Moreover, even the act of physically sexing the mosquitoes may affect them due to metabolic stress etc. In contrast, by using our sensors, we can allow unsexed pupae to hatch out and the adults fly in cages with order of magnitude larger volumes. In this way, we can automatically and noninvasively sex them to produce sex-specific daily flight activity plots.

## Conclusion and Future Work

In this work we have introduced a sensor/classification framework that allows the inexpensive and scalable classification of flying insects. We have shown experimentally that the accuracies achievable by our system are good enough to allow the development of commercial products and to be a useful tool for entomological research. To encourage the adoption and extension of our ideas, we are making all code, data, and sensor schematics freely available at the UCR Computational Entomology Page (Chen 2013). Moreover, within the limits of our budget, we will continue our practice of giving a complete system (as shown in Figure 2) to any research entomologist who requests one.

**Acknowledgements:** We would like to thank the Vodafone Americas Foundation, the Bill and Melinda Gates Foundation and São Paulo Research Foundation (FAPESP) for funding this research, and the many faculties from the Department of Entomology at UCR that offered advice and expertise.